\PassOptionsToPackage{dvipsnames}{xcolor}
\documentclass{adobe_research}
\usepackage[T1]{fontenc}
\usepackage{lmodern}
\usepackage{multirow}
\usepackage{booktabs}
\usepackage{soul}
\usepackage{pifont}
\usepackage[T1]{fontenc}

\usepackage{graphicx}
\usepackage{amsmath,amssymb}
\usepackage{tikz}
\usetikzlibrary{arrows.meta,positioning,calc,fit,shapes.geometric,decorations.pathreplacing}

\usepackage{makecell}
\usepackage{color, colortbl}
\usepackage{caption}
\usepackage{algorithm}

\usepackage{xspace}
\makeatletter
\DeclareRobustCommand\onedot{\futurelet\@let@token\@onedot}
\def\@onedot{\ifx\@let@token.\else.\null\fi\xspace}

\def\ie{\emph{i.e}\onedot} 
 
\def\etc{\emph{etc}\onedot}

\makeatother

\usepackage{newtxtext}

\usepackage{algpseudocode}
\definecolor{catgray}{gray}{0.92}

\definecolor{FutureOrange}{HTML}{EC866D}
\definecolor{teal}{HTML}{2F6F73}
\definecolor{orange}{HTML}{B86B2B}
\definecolor{purple}{HTML}{7A4D8F}
\newcommand{\ours}{\textcolor{FutureOrange}{\textbf{Wonder }}}

\title{\textcolor{FutureOrange}{\textbf{Wonder}}: Video World Model Done Better}

\author[*,1,2]{Jiacong Xu}
\author[1]{Hanwen Jiang}
\author[1]{Zhixin Shu}
\author[1]{Kalyan Sunkavalli}
\author[2]{\\Vishal M. Patel}
\author[*\dagger,1]{Yiqun Mei}
 \affiliation[1]{\textbf{Adobe Research}}
 \affiliation[2]{\textbf{Johns Hopkins University}}

\contribution[*]{Equal Contribution}
\contribution[\dagger]{Project Lead.}

\abstract{
We present \textcolor{FutureOrange}{\textbf{Wonder}}, a general-purpose video world model for real-time, camera-controllable world exploration. Given an image or a conditional video, \ours constructs a playable world where users can navigate interactively by moving the camera, discovering unseen regions, and revisiting previously observed areas in real time and over a long-term horizon. Achieving this capability requires a system-level co-design of control method, memory mechanism, and training strategy. We introduce a novel camera conditioning with a dense coordinate field whose renderings provide spatially aligned motion and orientation cues, allowing the model to interpret camera motion directly as visual evidence. To support fast and precise memory retrieval over a growing generation context, we propose an efficient sparse attention-based memory mechanism, enabling the model to selectively attend to a small set of relevant context tokens at inference time, regardless of actual context length. We further develop several techniques to rectify the self-forcing-style distillation pipeline, improving the student model's ability to respect control signals, as well as maintaining diverse generation modes and long-term memory from the teacher. Together, these components enable \ours to synthesize diverse, minute-scale videos at 16 FPS while preserving coherent geometry, appearance, and dynamics across long rollouts. Beyond image-to-video generation, \ours naturally supports video-conditioned generation, allowing existing dynamic scenes to be re-shot in real time.
}

\date{June 17th, 2026}

\adobedata[Project Page]{\url{https://wonder-world-model.github.io/}}

\begin{document}

\maketitle

\section{Introduction}
\label{sec:introduction}

Video generation is rapidly evolving from fixed-duration clip synthesis toward interactive visual world modeling, with potential applications in game creation~\citep{tang2025hunyuan2}, virtual production~\citep{rahimi2025generative}, and Robotics~\citep{xiao2025world}. In these settings, a model should construct a persistent visual environment from an initial condition, such as an image, a video, or a text prompt, and allow users or agents to explore it through control signals. This shift imposes requirements beyond visual fidelity alone. Interactive applications require accurate control following and low-latency response; simulation settings require long-horizon coherence; editing workflows require the model to preserve source content under control. Achieving these properties simultaneously remains challenging, because controllability, memory, and real-time efficiency often place conflicting demands on the model architecture and training pipeline.

Most existing methods ~\citep{wang2026matrix3, li2025hunyuangame, team2026dreamx} address only a subset of these requirements. Camera-controllable video generators provide accurate trajectory conditioning, but they are typically bidirectional, fixed-length, and computationally expensive, making them unsuitable for low-latency interaction or minute-scale rollout~\citep{bai2025recammaster, yu2025contextasmem}. Autoregressive video diffusion and distillation methods improve streaming efficiency by converting bidirectional models into causal few-step generators, but they often focus on video continuation and do not explicitly address precise camera following, persistent spatial memory, or consistent revisits~\citep{huang2025self, cui2025self, zhu2026causal}. Recent interactive world-model systems attempt to combine control, memory, and real-time generation, yet each component introduces new trade-offs: abstract pose or ray conditions can drift after distillation~\citep{zhu2026sana, team2025hunyuanworld}, reconstruction-based control is limited to observed views~\citep{yu2024viewcrafter, team2026inspatio}, dense historical attention increases latency with rollout length~\citep{gao2026advancing}, and compressed or sliding-window memory loses visual details~\citep{hong2025relic}. As a result, existing systems still struggle to achieve accurate control, faithful long-term memory, and stable real-time inference simultaneously.

In this report, we present \textcolor{FutureOrange}{\textbf{Wonder}}, a general-purpose video world model for real-time, camera-controllable world exploration. \textcolor{FutureOrange}{\textbf{Wonder}} follows the common teacher-to-student paradigm: adapting a bidirectional video diffusion model for camera control, converting it into a causal autoregressive generator, and distilling it into a few-step streaming model. Unlike prior systems that optimize these stages largely in isolation, \textcolor{FutureOrange}{\textbf{Wonder}} treats the pipeline as a coupled system. We jointly redesign the control representation, memory mechanism, and distillation strategy to preserve controllability, persistence, and efficiency from the teacher through the real-time student. As a result, \textcolor{FutureOrange}{\textbf{Wonder}} supports both image- and video-conditioned exploration with accurate camera following, coherent long-horizon memory, open-ended generation, and stable real-time latency. We specify our designs as follows:
\begin{enumerate}
    \item \textbf{Control.} We introduce a novel camera-control representation that exposes camera motion as rendered visual evidence rather than only as camera geometry. While pose~\citep{hong2025relic} or Plücker-ray~\citep{wang2026bullettime} provide per-image or pixel ray geometry, they still require the diffusion model to infer how those coordinates should translate into image motion. We instead render a synthetic 3D scaffold and environment map along the target camera trajectory, directly turning translation, rotation, and parallax into frame-aligned visual cues. Because the rendered scene is synthetic rather than reconstructed from the input, the control remains informative even beyond the observed view, enabling accurate and streamable camera control for open-ended image- and video-conditioned exploration. Moreover, because the rendered cues have explicit visual semantics, they are better preserved through few-step autoregressive distillation than abstract embeddings, reducing control drift in the real-time student.

    \item \textbf{Memory.} We develop a sparse full-fidelity memory mechanism for efficient long-horizon generation. Interactive world models require recent context for local motion continuity and distant context for persistent scene memory, but attending densely~\citep{gao2026advancing} to all historical KV states makes per-step latency grow with rollout length. Existing systems reduce this cost by compressing history~\citep{hong2025relic}, reconstructing context~\citep{huang2025memory}, using sliding windows~\citep{he2025matrixgame2}, or relying on geometry-based retrieval~\citep{team2026dreamx}, which trade off memory fidelity, revisit consistency, or retrieval robustness. Instead, \textcolor{FutureOrange}{\textbf{Wonder}} decouples memory storage from attention computation: it retains the growing historical KV cache in full fidelity, but uses lightweight pooled query-key summaries to select a constant-size set of relevant memory chunks at each rollout step. Only the selected chunks are attended with full-resolution keys and values, preserving faithful visual history for revisits while keeping active attention cost and per-step latency stable as the generated sequence grows. Furthermore, we introduce a co-designed training strategy, \ie sparse context forcing, that stably activates this sparse-memory capability in the student model, where the causal student can smoothly learn to generate content under the same retrieved-memory condition it will face at inference.

    \item \textbf{Distillation.} We redesign few-step autoregressive distillation around the goal of preserving teacher capabilities in the real-time student. The causal student must approximate a bidirectional many-step teacher with far fewer denoising steps, which leads to both reduced generation diversity and accumulated camera drift. We address these two losses with a timestep-wise mixture-of-students, which increases effective student capacity without changing the streaming interface, and a camera-aware adversarial regularizer, which provides low-frequency control supervision during distillation. Together, they allow the few-step student to retain visual quality, diversity, and camera-following accuracy during long autoregressive rollouts.
\end{enumerate}

\begin{figure}
    \centering
    \includegraphics[width=1\linewidth]{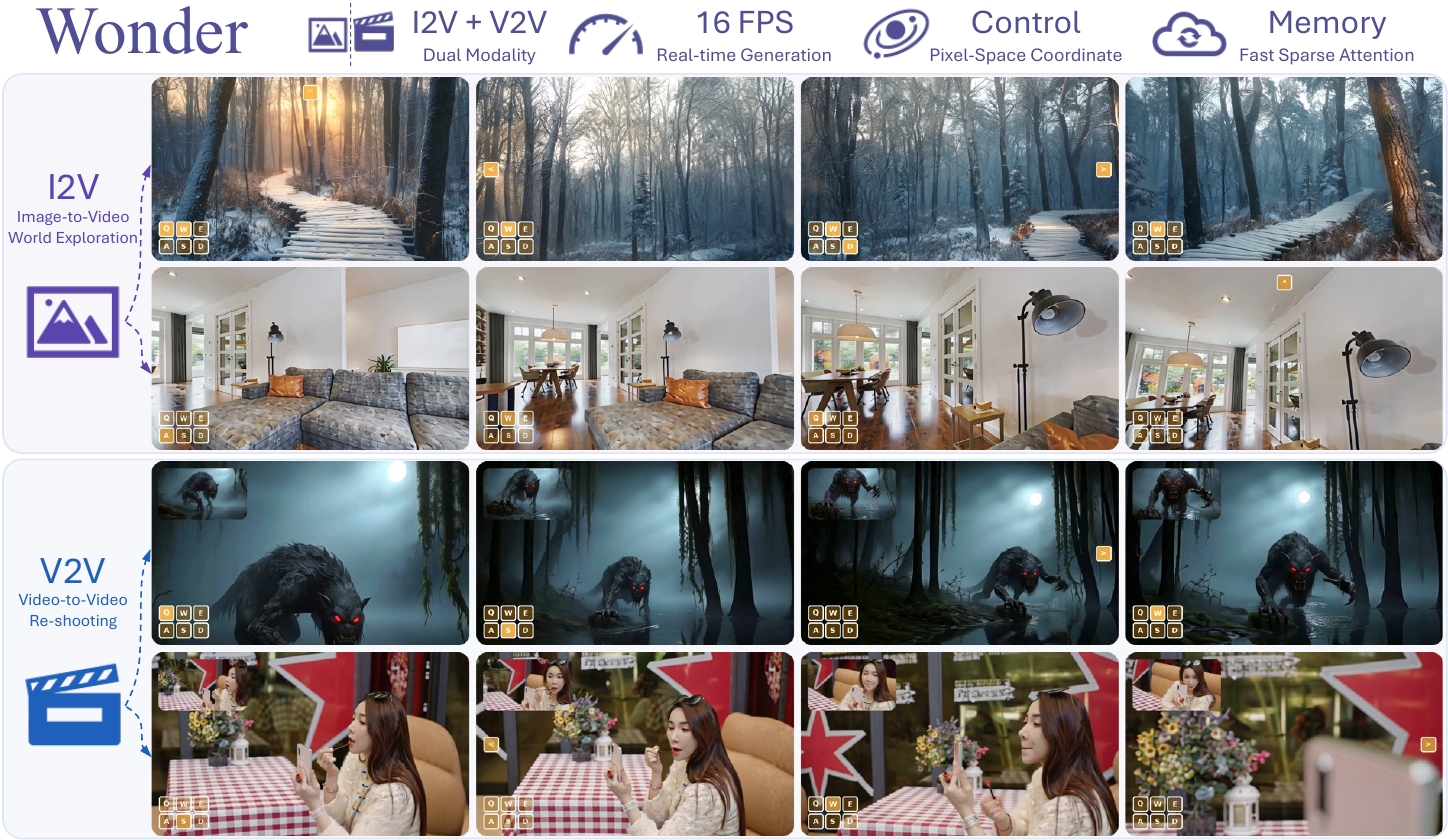}
    \caption{
    \textbf{\ours enables interactive world exploration from images and videos.}
    Given a single image, \ours turns it into a navigable world where users can move the camera to explore both visible regions and plausible unseen areas beyond the input view. Given a video, \ours constructs a dynamic world that preserves the source motion while allowing users to freely change the camera trajectory, enabling camera-controllable 4D video exploration.
    }
    \label{fig:teaser}
\end{figure}

Experiments in both image- and video-conditioned settings show that \ours improves visual quality and camera-following accuracy over recent streaming world models. The model supports exploration beyond observed views, preserves dynamic content when re-rendering source videos, and maintains coherent revisits over long trajectories. It generates minute-scale rollouts at 16 FPS with stable latency as the history grows.

\section{Related Work}
\label{sec:relatedwork}

\paragraph{Video Diffusion Models.}
Modern text- and image-to-video generators are commonly built upon diffusion transformers operating in the latent space of a 3D VAE. Open-source DiT/MMDiT-style backbones, such as CogVideoX~\citep{yang2025cogvideox}, HunyuanVideo~\citep{kong2024hunyuanvideo}, Wan~\citep{wan2025wan}, and LTX-Video~\citep{hacohen2026ltx}, can synthesize multi-second, high-fidelity video clips and have become the de facto pretrained foundations for downstream video generation tasks. However, these models are bidirectional and denoise a fixed-length chunk in a single shot, making them poorly suited for interactive control and unbounded rollout.

\vspace{-0.1in}
\paragraph{Autoregressive Video Generation.}
A second line of work transforms bidirectional video diffusion models into causal, autoregressive generators. Diffusion Forcing~\citep{chen2024dforcing} unifies next-token prediction with diffusion by assigning independent noise levels to individual tokens. CausVid~\citep{yin2025causvid} distills a bidirectional teacher into a few-step causal student through distribution matching distillation~\citep{yin2024dmd1}. Self-Forcing~\citep{huang2025self} further narrows the training--inference gap by rolling out the student on its own predictions during training, while Rolling Forcing~\citep{liu2025rolling} and Causal Forcing~\citep{zhu2026causal} extend this paradigm to long-horizon, real-time streaming generation. Together, these methods establish the low-latency, KV-cache-enabled autoregressive foundation that makes interactive video world models practical.

\vspace{-0.1in}
\paragraph{Camera-controllable Video Generation.}
Pose-conditioned video diffusion has become a standard approach for trajectory-controlled video synthesis. MotionCtrl~\citep{wang2024motionctrl} injects 6-DoF camera extrinsics into video diffusion models, CameraCtrl~\citep{he2025cameractrl} introduces Pl\"ucker embeddings as a plug-in adapter, and ViewCrafter~\citep{yu2024viewcrafter} conditions generation on point-cloud renderings to improve geometric consistency. ReCamMaster~\citep{bai2025recammaster} formulates camera control as video-to-video re-rendering along novel trajectories, while subsequent works~\citep{wang2026bullettime,huang2025spacetimepilot} further disentangle camera motion from scene dynamics and spatiotemporal correlations. These camera-conditioning designs have inspired recent advances in video world models.

\vspace{-0.1in}
\paragraph{Interactive Video World Models.}
Most closely related to our work are neural simulators that respond to user actions. Genie~3~\citep{genie3} generates general-purpose, text-promptable interactive worlds, representing the current closed-source frontier. In parallel, a rapidly growing open-source line of work, including Matrix-Game and its real-time/streaming successors~\citep{zhang2025matrixgame1,he2025matrixgame2,wang2026matrix3}, MineWorld~\citep{guo2025mineworld}, Hunyuan-GameCraft~\citep{li2025hunyuangame} and its instruction-following extension Hunyuan-GameCraft-2~\citep{tang2025hunyuan2}, Yume~\citep{mao2025yume}, and RELIC~\citep{hong2025relic}, has explored real-time streaming, long-horizon memory, and richer action interfaces. More recently, LingBot-World~\citep{gao2026advancing} and DreamX-World~\citep{team2026dreamx} achieve real-time streaming with minute-level context across photorealistic, game-style, and cartoon domains. Closest to our setting, Inspatio-World~\citep{team2026inspatio} provides both I2V and V2V baselines for real-time dynamic world modeling and re-rendering. These prior works advance video world modeling from different perspectives, but still face persistent challenges such as camera drift, memory loss, and degraded visual quality during long-horizon rollout. In contrast, \ours adopts a system-level co-design across control representation, memory mechanism, and distillation strategy, effectively alleviating these issues and enabling more stable interactive world exploration.

\section{Training Data}
\label{sec:data}

\begin{figure}[t]
    \centering
    \includegraphics[width=1\linewidth]{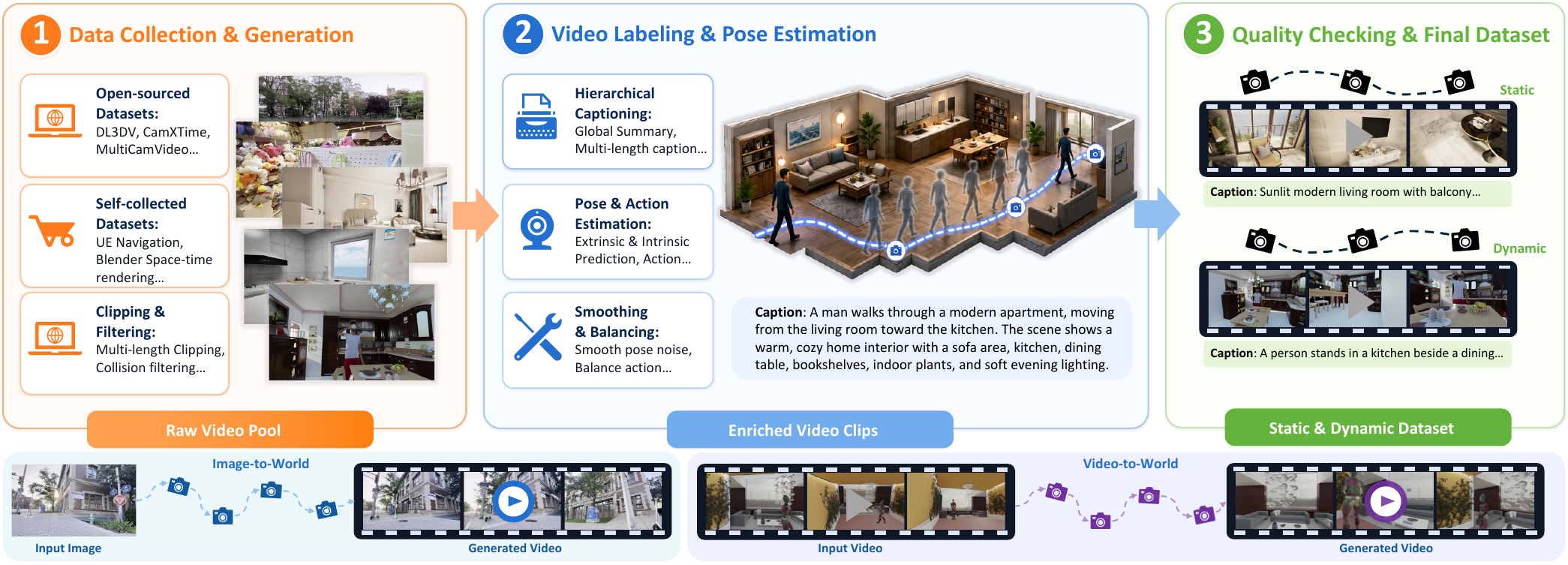}
\caption{\textbf{Data curation pipeline.}
We construct a unified data engine for image-to-world (I2W) and video-to-world (V2W) training. We first collect videos from public datasets and self-rendered UE/Blender environments, then perform multi-duration clipping, collision filtering, and rule-based quality screening to build a raw video pool. Each clip is further enriched with hierarchical VLM captions and camera poses estimated by Depth Anything 3 (DA3), from which we derive discrete camera actions. To improve trajectory quality and action coverage, we apply Gaussian pose smoothing and temporal augmentations such as reverse playback and speed resampling. A final VLM-based quality check removes visually degraded clips and sequences with unnatural camera motion, yielding the curated static and dynamic datasets used for controllable world-model training.}
    \label{fig:data}
\end{figure}

Training a controllable world model requires large-scale videos with diverse visual content, accurate camera motion, and sufficiently long temporal horizons. Compared with short video generation, long-horizon world modeling requires the model to enforce memory, preserving scene geometry, appearance, and dynamics over extended rollouts. Therefore, we construct a unified data curation pipeline, as shown in Fig.~\ref{fig:data}, to collect, annotate, and filter video clips.

For I2V training, we first leverage DL3DV \citep{ling2024dl3dv} as a major source of real-world navigation videos. It contains diverse scenes and naturally captured camera trajectories, making it useful for learning realistic geometry-aware motion. However, the camera movements in such real-world videos are often relatively simple and smooth, while interactive users may perform more challenging controls such as sharp turns, lateral movements, backward motion, and compound camera actions. To better cover these hard cases, we additionally build a large-scale synthetic I2V dataset with Unreal Engine (UE). By rendering diverse indoor and outdoor environments with user-controlled-like trajectories, the synthetic data provides richer supervision for robust camera-controllable world exploration.

For V2V training, the data requirement is more challenging because the model needs paired videos that share the same underlying dynamic scene but differ in camera trajectory or temporal evolution. Although open-source datasets such as MultiCamVideo \citep{bai2025recammaster} and CamXTime \citep{huang2025spacetimepilot} provide useful paired videos, their clips are typically short, often around 5 seconds, which is insufficient for long-horizon world modeling. We therefore render a large amount of long paired V2V data using Blender. The rendered pairs include standard paired trajectories, speed-varied sequences, and bullet-time videos, where camera motion and scene time are partially decoupled. These data help the model learn the distinction between spatial viewpoint changes and temporal scene dynamics, which is essential for controllable V2V generation.

After constructing the raw video pool, we apply a unified annotation and filtering pipeline. We first clip videos into multiple durations, including 5s, 10s, and 20s, to support progressive training from short to long horizons. Rule-based filters are used to remove corrupted videos, black frames, failed renderings, and trajectories with collision or clearly invalid motion. We then use a VLM~\citep{bai2025qwen3} to generate hierarchical captions~\citep{gao2026advancing} for each long clip, consisting of a global video-level summary and fine-grained captions for its short temporal sub-clips. For metric camera supervision, we use Depth Anything 3 (DA3) \citep{lin2025depth} to estimate camera intrinsics and extrinsics for both real and synthetic videos. The recovered trajectories are converted into discrete camera actions. Since estimated poses may contain high-frequency jitter, we apply Gaussian smoothing before action discretization, which suppresses noisy pose fluctuations while preserving the overall camera path. We further use temporal augmentations such as reverse playback and speed resampling to balance the action distribution. Finally, an additional VLM-based quality checking stage removes videos with low visual quality, unnatural camera motion, or inconsistent temporal dynamics, resulting in the final curated static and dynamic datasets for training.

\section{\ours\ World Model}
\label{sec:method}
\ours generates explorable videos that follow a user-streamed camera trajectory. It supports two inference modes. In image-to-video mode, a single input frame initializes an explorable world. In video-to-video mode, a source video anchors the appearance and motion of an existing event, while the target camera trajectory specifies a new viewpoint of that event. By unifying both modes within a single model, \ours enables multimodal video world modeling.

Following the common framework of recent works~\citep{hong2025relic, gao2026advancing}, we adopt a two-stage pipeline that distills a few-step autoregressive video diffusion model from a bidirectional video diffusion teacher, with both models conditioned on camera-control signals. While this framework has shown promising early results, achieving high-quality, low-latency, memory-aware long-horizon rollout remains an open challenge. Existing approaches often fail to satisfy these requirements simultaneously: some~\citep{gao2026advancing, team2025hunyuanworld} struggle to faithfully follow camera instructions; many~\citep{zhu2026sana, he2025matrixgame2, li2025hunyuangame} lack an explicit memory mechanism or rely on lossy scene summaries~\citep{hong2025relic}; and methods that preserve memory by retaining the full historical KV cache~\citep{gao2026advancing} incur latency that grows linearly with context length. These limitations motivate us to systematically redesign the pipeline, including the control representation, memory retrieval mechanism, and autoregressive training strategy.

\begin{figure}[t]
    \centering
    \resizebox{\linewidth}{!}{%
    \begin{tikzpicture}[
        node distance=0.7cm,
        box/.style={draw=black!35, rounded corners=2pt, fill=white, minimum width=3.1cm, minimum height=1.0cm, align=center, font=\small},
        stage/.style={box, minimum width=3.5cm, minimum height=1.25cm, font=\bfseries\small},
        sub/.style={draw=black!25, rounded corners=2pt, fill=black!3, align=center, font=\scriptsize, minimum width=3.2cm, minimum height=0.55cm},
        arrow/.style={-{Latex[length=2.3mm]}, line width=0.7pt, draw=black!65}
    ]
        \node[stage, fill=teal!10] (s1) {Stage 1\\Bidirectional teacher};
        \node[stage, fill=orange!13, right=0.55cm of s1] (s2) {Stage 2\\Sparse ODE init};
        \node[stage, fill=purple!11, right=0.55cm of s2] (s3) {Stage 3\\Few-step AR student};
        \node[stage, fill=FutureOrange!15, right=0.55cm of s3] (s4) {Streaming\\Inference};

        \draw[arrow] (s1) -- (s2);
        \draw[arrow] (s2) -- (s3);
        \draw[arrow] (s3) -- (s4);

        \node[sub, below=0.55cm of s1] (s1a) {Pixel-Space Coordinate \\\textit{Camera Control}};
        \node[sub, below=0.18cm of s1a] (s1b) {Multi-Horizon Training \\\textit{I2V+V2V, 5s to 20s}};

        \node[sub, below=0.55cm of s2] (s2a) {ODE Latent Pairs \\\textit{Coarse Adaptation}};
        \node[sub, below=0.18cm of s2a] (s2b) {Sparse Context Forcing \\\textit{Memory Retrieval}};

        \node[sub, below=0.55cm of s3] (s3a) {Mixture of Students \\\textit{Generation Fidelity}};
        \node[sub, below=0.18cm of s3a] (s3b) {GAN Control Regularization \\\textit{Camera Supervision}};

        \node[sub, below=0.55cm of s4] (s4a) {Runtime Optimization \\\textit{Compilation, Caching \etc}};
        \node[sub, below=0.18cm of s4a] (s4b) {Multi-GPU Parallelism\\\textit{Workload Distribution}};
    \end{tikzpicture}
    }
    \caption{\textbf{Training and deployment pipeline.} \ours\ first learns high-quality camera control in a bidirectional teacher, then converts the model to a sparse-context causal student, distills it into a few-step autoregressive generator, and finally deploys it with a resident streaming runtime.}
    \label{fig:model_pipeline}
\end{figure}
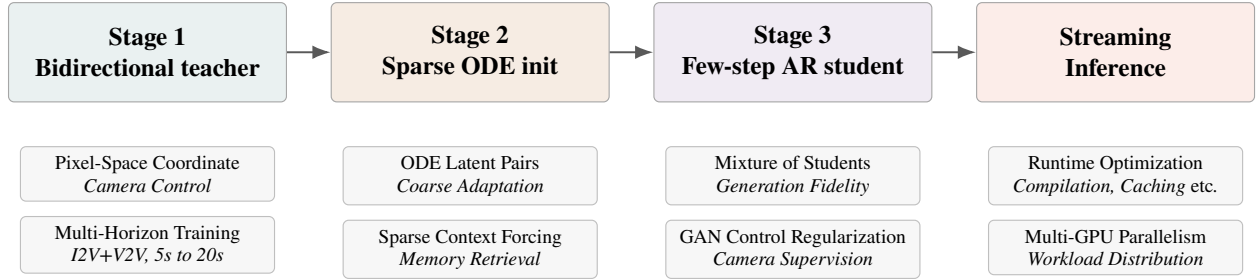

We highlight several key innovations in our framework, as shown in Figure \ref{fig:model_pipeline}. First, we redesign the bidirectional video diffusion teacher to condition on a dense pixel-space camera representation, enabling camera motion to be interpreted as a pixel-aligned visual signal and yielding strong action-following behavior (Sec.~\ref{sec:lattice}). We then distill this teacher into a few-step autoregressive video diffusion model for low-latency interactive streaming. To support long-horizon generation without increasing inference cost, we introduce a lossless yet efficient memory mechanism that selectively retrieves a compact set of faithful historical video tokens from the KV cache. This mechanism is paired with a training strategy that induces emergent memory retrieval during autoregressive rollout (Sec. ~\ref{sec:sparse}). We further improve the self-forcing training paradigm with two techniques that mitigate quality degradation and camera drift during distillation (Sec.~\ref{sec:mos} \& \ref{sec:gan}), substantially improving both training stability and controllability. Finally, we apply runtime optimizations that enable real-time inference and minute-scale generation.

\subsection{Stage 1: Camera-Conditioned Teacher for Long Video Generation}
\label{subsec:stage1}
Training a strong bidirectional teacher with precise camera following and multimodal input support is essential in our setting, as it serves both as the initialization point for the autoregressive student and as the supervision signal for distillation. We next describe the techniques used to obtain this bidirectional teacher.

\subsubsection{Pixel-Space Coordinate Field}
\label{sec:lattice}

\begin{figure}
    \centering
    \includegraphics[width=1\linewidth]{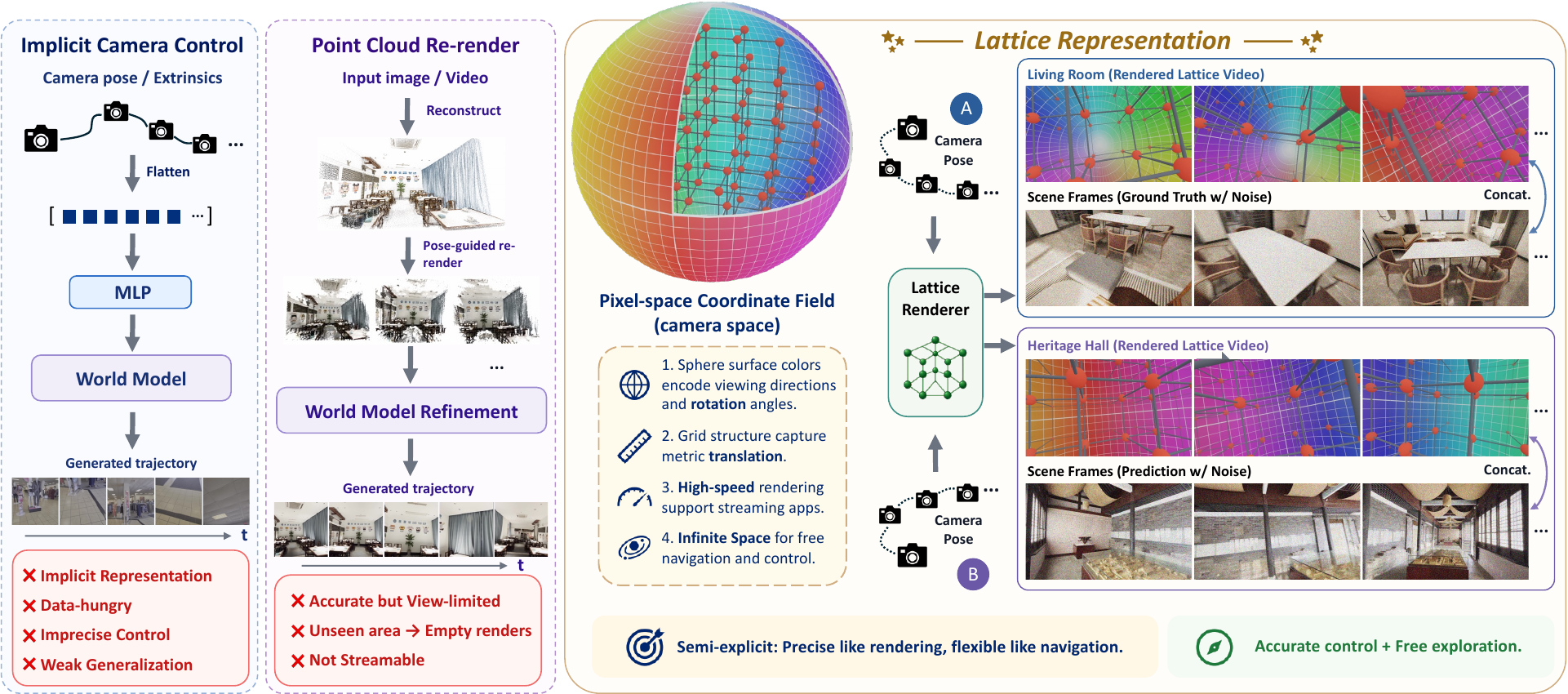}
    \caption{\textbf{Camera control using Pixel-Space Coordinate Field.}
Prior camera-control methods are either implicit, encoding camera poses with an MLP~\citep{bai2025recammaster} or RoPE~\citep{wang2026bullettime}, or explicit, re-rendering reconstructed point clouds~\citep{team2026inspatio}. The former is data-hungry and often imprecise, while the latter is accurate but view-limited and fails when the camera explores unseen regions. 
Our lattice representation combines the strengths of both. We construct a pixel-space coordinate field with a colored spherical environment map at infinity and a dense 3D lattice scaffold inside it. A lightweight renderer converts arbitrary camera trajectories into pixel-aligned lattice videos that encode viewing direction and metric translation. Concatenating these renders with noisy scene inputs provides precise, streamable, and exploration-friendly camera control for world modeling.}
    \label{fig:lattice}
\end{figure}

As shown in Figure~\ref{fig:lattice}, existing camera-control methods can be broadly grouped into two categories: implicit and explicit. Implicit methods typically represent camera motion as a sequence of low-dimensional vectors derived from camera extrinsics~\citep{gao2026advancing, bai2025recammaster, hong2025relic}. These vectors are projected into high-dimensional embeddings and injected token-wise into the diffusion transformer, where all spatial tokens within the same latent frame share an identical camera condition. Although compact, such global numerical representations provide only coarse motion cues and are therefore difficult for the model to associate with subtle, local, or high-frequency pixel-space changes, even with large-scale training. Ray-based representations, as well as their combinations with RoPE~\citep{wang2026bullettime, team2026dreamx}, offer a more geometric alternative, but still suffer from ambiguities in scale, origin, and depth. These limitations become even more pronounced during distillation: commonly used objectives such as DMD~\citep{yin2024dmd1} provide no direct supervision for camera following, which often results in camera drift and weak generalization in distilled models for diverse camera movements.

In contrast, explicit methods~\citep{team2026inspatio, yu2024viewcrafter} rely on offline pose or geometry estimation models to reconstruct a point cloud from the input image or video, which is then re-rendered from user-specified viewpoints. Such representations provide a strong geometric anchor and reduce the role of the world model to refinement or inpainting. However, they also introduce several practical limitations. First, their effectiveness heavily depends on the quality and robustness of external pose, depth, or reconstruction models. Second, the resulting control signal is inherently view-limited: when the camera moves beyond the regions observed in the input, the re-rendered point cloud often becomes empty or black, providing little useful camera guidance and causing control failures. Finally, accurate reconstruction typically requires buffering multiple frames and running additional external modules, which introduces latency and complicates both training and inference pipelines. These requirements make explicit re-rendering difficult to deploy in low-latency streaming applications.

To overcome these limitations, we introduce a pixel-space camera representation that exposes camera motion as visual evidence. Specifically, we construct a synthetic camera space and render it from the target camera trajectory, producing conditioning frames that are spatially aligned with the video to be generated. The synthetic scene consists of two complementary components. First, we place a dense 3D scaffold in the camera space, which provides an explicit spatial parameterization of the environment and alleviates the depth ambiguity. Its 2D projections deform under camera translation and induce parallax across frames, allowing the model to infer metric camera translation and align the rendered structure with the conditional image or video. Second, we place a colored environment map at infinity, whose appearance varies with camera orientation and thus provides a clear cue for rotation. Together, the scaffold and the environment map convert translation and rotation into dense, frame-aligned visual signals that are easy for the video diffusion model to interpret. We implement this representation with a lightweight OpenGL renderer, which synthesizes conditioning frames at 150 FPS with negligible overhead. The rendered frames are encoded by the VAE and concatenated channel-wise with the target noisy latents before being processed by the DiT. Empirically, this representation substantially improves camera controllability and stabilizes the distillation process.

\subsubsection{\textbf{Base Architecture}}
We build our framework on Wan2.1-I2V-14B~\citep{wan2025wan}, a bidirectional video diffusion model originally developed for short-horizon image-to-video generation. Wan consists of two main components: a spatio-temporal VAE and a diffusion transformer backbone. The VAE maps input videos into compact latent sequences with a 3D causal encoder-decoder, downsampling the spatial resolution by a factor of 8 and the temporal resolution by a factor of 4. Its causal design and feature-caching mechanism provide a natural and efficient interface for streaming video generation in our system. The diffusion backbone operates on patchified video latents. After a stack of transformer blocks, the latent tokens are mapped back to video latents through corresponding unpatchifying layers. Text prompts are encoded by umT5~\citep{Chung2023UniMaxFA} and injected via cross-attention, while diffusion timesteps are embedded by a shared MLP and used to modulate the transformer blocks. This architecture provides a strong bidirectional teacher initialization, which we further adapt for camera-controllable world exploration.

To support both image- and video-conditioned generation within a unified framework, we represent the model input as an optional source segment followed by a target segment. For video-conditioned generation, the source segment corresponds to the conditioning video and is provided as clean latents. For image-conditioned generation, there is no source segment; instead, the first frame of the target video is directly provided as a clean, noise-free latent. Noise is applied only to the target frames that need to be generated, while the conditioning latents remain clean throughout both training and inference. Importantly, the binary mask is used to distinguish source and target segments. The mask value is set to $0$ for the source video and $1$ for the target video. Similarly, the encoded camera-condition latents are zero-padded only for the source segment, and are provided for the entire target trajectory. Formally, let $\mathbf{z}^{s}$ denote the optional source-video latents, and let the target latents be decomposed as $\mathbf{z}^{t}=[\mathbf{z}^{a},\mathbf{z}^{p}]_{T}$, where $\mathbf{z}^{a}$ denotes the clean target anchor frame and $\mathbf{z}^{p}$ denotes the target frames to be generated. The model input at diffusion timestep $\tau$ is constructed as
\begin{equation}
\mathbf{u}_{\tau}
=
\left[
\left[\mathbf{z}^{s}, \mathbf{z}^{a}, \alpha_{\tau}\mathbf{z}^{p}+\sigma_{\tau}\boldsymbol{\epsilon}\right]_{T},
\left[\mathbf{0}^{s}_{m}, \mathbf{1}^{a}_{m}, \mathbf{1}^{p}_{m}\right]_{T},
\left[\mathbf{0}^{s}_{c}, \mathbf{z}^{c,a}, \mathbf{z}^{c,p}\right]_{T}
\right]_{C}.
\end{equation}
Here, $\boldsymbol{\epsilon}\sim\mathcal{N}(\mathbf{0},\mathbf{I})$ is Gaussian noise, and $[\cdot]_{T}$ and $[\cdot]_{C}$ indicate concatenation along the temporal and channel dimensions. For positional encoding, we mirror the RoPE coordinates of the target video onto the source video, and add separate learnable positional embeddings to help the model distinguish source and target tokens.

\subsubsection{Multi-task Multi-Horizon Training}
We train a single student model that supports both image-to-video and video-to-video generation, while being able to roll out to long horizons. Following the progressive-horizon distillation strategy of~\citep{hong2025relic}, we gradually extend the training horizon instead of directly training on the longest sequences. Concretely, we first post-train the base model on 5-second clips with both I2V and V2V tasks jointly enabled. We then extend the horizon to 10 seconds and finally to 20 seconds, where the RoPE positions are extrapolated with YaRN~\citep{peng2023yarn} to support longer temporal contexts.

A direct multi-task extension to long horizons is computationally inefficient, since V2V samples contain both source and target videos and therefore incur substantially higher memory and compute cost. We therefore adopt a mixed-length training schedule in the long-horizon stages. For most training iterations, we keep V2V samples at the 5-second horizon while training I2V samples at the current long horizon, i.e., 10 or 20 seconds. This design is based on the observation that long-horizon camera following can be effectively learned from I2V data, while V2V mainly teaches source-target correspondence and can be learned from shorter paired clips. The remaining iterations still expose the model to full-horizon multi-task samples to preserve compatibility across tasks. This strategy maximizes training throughput while enabling the model to generalize to long-horizon V2V dynamic world exploration.

\subsection{Stage 2: Fast Autoregressive Student for Real-Time Streaming}
The camera-conditioned teacher provides accurate control, but it is not directly interactive: it is bidirectional, relies on dense attention, and requires many denoising steps, making live control prohibitively expensive. Following recent video world model pipelines~\citep{gao2026advancing,team2025hunyuanworld,hong2025relic}, we adapt and distill this teacher into an autoregressive few-step student for low-latency streaming generation. We base our distillation framework on self-forcing~\citep{huang2025self}, a rollout-based distillation paradigm designed to reduce exposure bias in long-horizon generation. Specifically, the student is initialized from the teacher and first trained with ODE initialization, where it regresses precomputed teacher ODE trajectories at four denoising timesteps. We then apply DMD~\citep{yin2024dmd1} on the student's own autoregressive rollouts, minimizing the reverse KL divergence between the diffused real-data distribution and the student's self-induced generation distribution over sampled timesteps. 
 
\begin{figure}[t]
    \centering
    \includegraphics[width=1\linewidth]{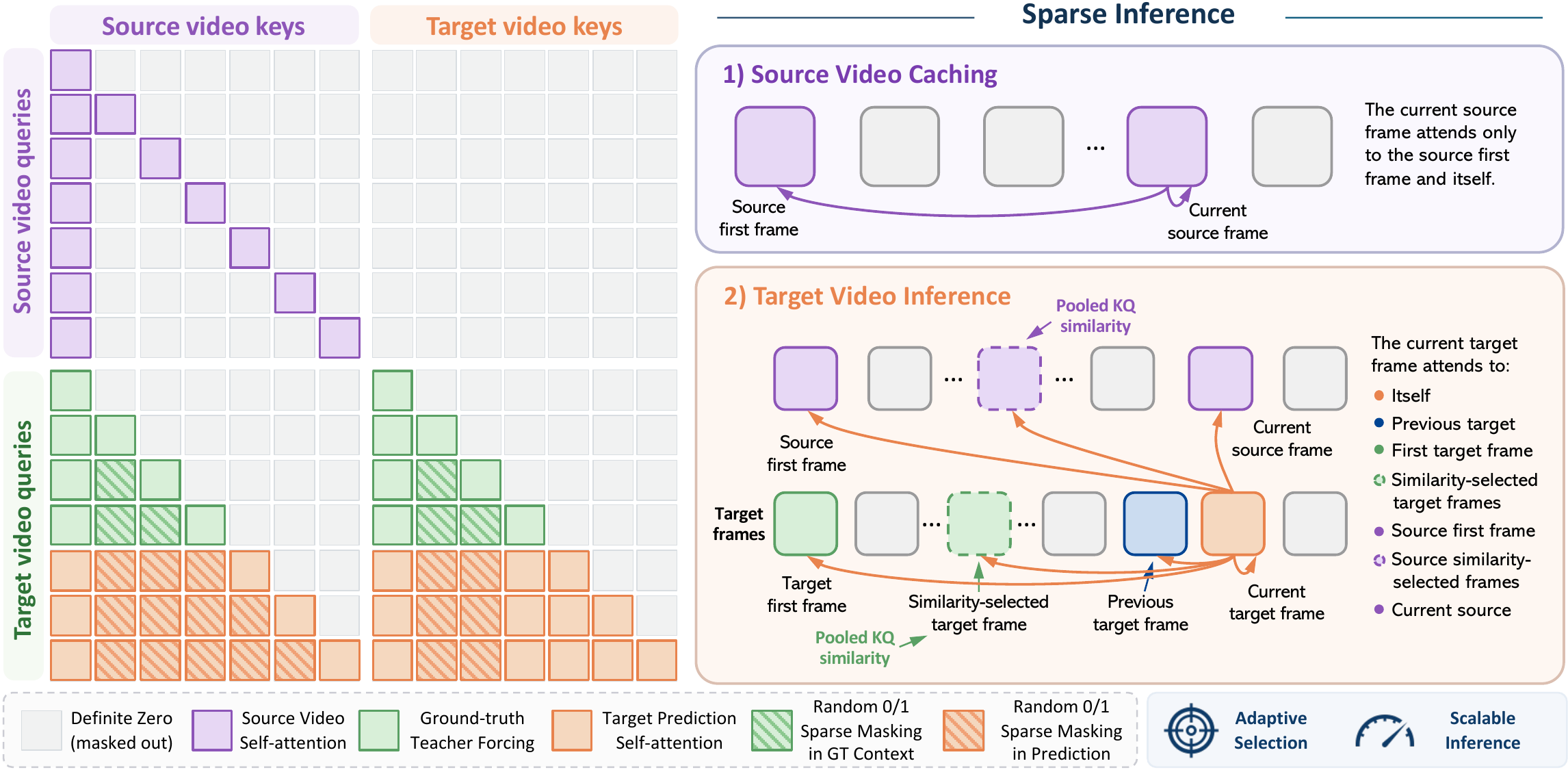}
    \caption{\textbf{Sparse context forcing and sparse inference.}
Left: during ODE initialization, we train the causal student with a unified attention layout for I2V and V2V. Source-video tokens attend only to the first source frame and themselves, while target-video tokens combine teacher-forced clean context and diffusion-forced predicted context. To prepare the student for sparse-memory inference, we randomly drop eligible non-local context edges in both ground-truth and prediction regions, while preserving required anchors such as self-attention, first-frame anchors, and recent context. 
Right: at inference time, the source video is cached as a reference stream, and each target chunk attends to a constant-size set of memories, including itself, recent target chunks, first-frame anchors, current/source references, and additional source/target chunks selected by pooled query-key similarity.
}
    \label{fig:sparse}
\end{figure}

\subsubsection{Sparse Context Forcing}
\label{sec:sparse}
During autoregressive rollout, the student continuously writes generated chunks into the KV cache. As the rollout grows longer, the cache size increases accordingly, causing the generation speed to decrease approximately linearly with the context length~\citep{gao2026advancing}. However, world modeling requires persistent memory to support consistent revisits to previously observed regions. Therefore, the cache cannot be simply discarded with a sliding-window strategy~\citep{zhu2026sana,li2025hunyuangame,team2026dreamx}. Existing solutions are still limited. One approach compresses distant historical frames in the latent space~\citep{hong2025relic}, but aggressive compression often leads to noticeable degradation in visual quality and weakens long-term consistency. Another line of work retrieves relevant memory by estimating the field-of-view overlap between the current viewpoint and historical viewpoints~\citep{yu2025contextasmem,team2026dreamx}. While more adaptive, such geometry-based heuristics are brittle in practice: they depend on reliable camera geometry and can fail in corner cases, such as revisiting the same location from opposite viewpoints.

Video diffusion models naturally learn rich cross-frame correspondences~\citep{xu2025freevis}. Instead of compressing or discarding historical KV states, we therefore keep the entire KV cache in full fidelity and let the model retrieve relevant chunks through content-aware sparse attention. As shown in Figure \ref{fig:sparse}, each historical chunk stores its dense keys and values for attention, while only a lightweight pooled key summary is used for retrieval. At each rollout step, the current query summary is compared with the summaries of historical chunks to select a small set of active memory chunks. We always keep the initial chunk and the most recent $r$ chunks active, where the former serves as an attention sink, and the latter preserves short-term motion continuity. 

Let $\mathcal{I}_c$ denote the token index set of cache chunk $c$, and let
$\mathbf{K}_{\mathcal{I}_c}$ and $\mathbf{V}_{\mathcal{I}_c}$ be the corresponding full-resolution keys and values. We summarize each historical chunk by a pooled key $\bar{\mathbf{K}}_c$, and summarize the current query chunk as $\bar{\mathbf{Q}}_t$. The active memory set $\mathcal{A}_t$ is then selected by combining the initial chunk, the most recent $r$ chunks, and the top-$k$ middle-history chunks with the highest query-key similarity:
\begin{equation}
\begin{aligned}
\mathcal{A}_t
&=
\{0\}\cup\mathcal{N}_{r}(t)
\cup
\operatorname{TopK}_{k}
\left\{
\operatorname{sim}\!\left(\bar{\mathbf{Q}}_t,\bar{\mathbf{K}}_c\right)
\,\middle|\,
c\in\mathcal{M}(t)
\right\}, \\
\mathbf{K}^{\mathrm{ctx}}_t
&=
\left[\mathbf{K}_{\mathcal{I}_c}\right]_{c\in\mathcal{A}_t},
\qquad
\mathbf{V}^{\mathrm{ctx}}_t
=
\left[\mathbf{V}_{\mathcal{I}_c}\right]_{c\in\mathcal{A}_t}.
\end{aligned}
\label{eq:sparse_retrieval}
\end{equation}
Here, $\mathcal{N}_{r}(t)$ denotes the most recent $r$ chunks before the current chunk, and $\mathcal{M}(t)$ denotes the middle-history chunks excluding the initial and recent chunks. $\operatorname{TopK}_{k}$ returns the indices of the $k$ chunks with the highest similarity scores, and $[\cdot]_{c\in\mathcal{A}_t}$ denotes concatenation over the selected chunks in their original temporal order. In practice, we set $r=2$ to preserve local motion smoothness. Since retrieval is performed only on lightweight pooled summaries, the selection overhead is negligible. Meanwhile, the selected chunks are still attended with their full-resolution keys and values, allowing the model to preserve faithful visual memory without latent compression. Under a fixed top-$k$ budget, the active attention context remains constant regardless of rollout length, enabling long-horizon generation with stable latency.

Directly applying sparse attention during distribution-level distillation is unstable, since the student has not yet learned to generate under incomplete historical context. We therefore introduce \emph{sparse context forcing} during ODE initialization, so that the student is warmed up with the same memory constraint it will encounter at inference time. Our ODE initialization combines teacher forcing~\citep{gao2024ca2,hu2024acdit} and diffusion forcing~\citep{chen2024dforcing} under a causal attention layout. Teacher-forced context provides clean target frames to simulate low-noise historical context, while diffusion-forced context exposes the student to previous noisy target frames, reducing the train--test mismatch during autoregressive rollout~\citep{hong2025relic}. For video-to-video training, the source video is additionally cached as a reference stream, where each source frame attends only to itself and the first source frame. For image-to-video training, the layout naturally degenerates to the target-only case.

To make the student robust to sparse memory retrieval, we randomly drop eligible historical context edges during ODE initialization. Required edges, such as self-attention, the first-frame anchors, and the most recent context frames, are always preserved to maintain identity and local motion continuity. Non-local historical edges are treated as optional memory and are randomly kept on each layer and each forward pass. Formally, let $a(i,j)$ denote the base causal relation between query chunk $i$ and key/value chunk $j$. We define $\mathcal{R}$ as the set of required relations and $\mathcal{O}$ as the set of optional historical-context relations. The attention mask at layer $\ell$ is sampled as
\begin{equation}
M^{(\ell)}_{ij}
=
\mathbf{1}\!\left[a(i,j)\in\mathcal{R}\right]
+
b^{(\ell)}_{ij}\,
\mathbf{1}\!\left[a(i,j)\in\mathcal{O}\right],
\qquad
b^{(\ell)}_{ij}\sim\operatorname{Bernoulli}(1-p_{\ell}).
\label{eq:sparse_context_mask}
\end{equation}
Here, $p_{\ell}$ controls the sparsity level of optional memory edges. We use a stronger drop rate for longer target horizons, encouraging the model to rely on a smaller set of informative memories instead of dense access to the full history. This training strategy aligns the ODE-initialized student with our sparse inference pattern: the model learns to preserve quality and camera consistency even when only a limited subset of historical chunks is available.

\subsubsection{Mixture of Students}
\label{sec:mos}
After ODE initialization, the student can already recover coarse scene structure and layout, but still struggles to synthesize fine-grained local details. We therefore further optimize the student with DMD~\citep{yin2024dmd1} to better align its rollout distribution with the data distribution. As illustrated in Figure~\ref{fig:method}, the few-step causal student autoregressively generates full video rollouts conditioned on previously generated clean chunks. These rollouts are then perturbed with noise and fed into the bidirectional teacher and critic for score estimation. The resulting score difference provides the gradient signal for updating the student generator.

Self-forcing~\citep{huang2025self} reduces exposure bias by training the student on its own autoregressive rollouts. However, the underlying distillation problem remains highly challenging: the student is required to approximate a strong bidirectional, many-step teacher with a causal architecture and only a few denoising steps. These architectural and sampling constraints significantly reduce the effective modeling capacity of a single student, making it difficult to capture the diverse modes of the teacher distribution. In practice, this often leads to mode shrinkage, weakened local realism, and degraded fine details during long-horizon rollout. This motivates us to introduce a mixture-of-students design, which increases the student-side capacity while preserving the same autoregressive few-step inference interface.

Motivated by the capacity bottleneck discussed above, we adopt a timestep-wise mixture-of-students design, analogous to the denoising-stage MoE in Wan2.2~\citep{wan2025wan}. Instead of using a single generator for all few-step updates, our 4-step sampler uses three student generators: $G_1$ for the first step, $G_2$ for the second step, and $G_3$ for the last two steps:
\begin{equation}
\hat{\mathbf{x}}_{0}
=
G_3^{(4)}\!\circ G_3^{(3)}\!\circ G_2^{(2)}\!\circ G_1^{(1)}(\mathbf{x}_{\tau_1}),
\end{equation}
where the superscript denotes the denoising step. This design allocates capacity according to the role of each denoising stage: $G_1$ focuses on coarse structure formation, $G_2$ performs structure refinement, and $G_3$ is responsible for detail refinement and final reconstruction.

To preserve streaming efficiency, we do not maintain separate KV caches for different students. Instead, all generators share the autoregressive interface and reuse the KV cache produced by $G_3$. Therefore, the mixture-of-students increases the effective student-side capacity without changing the number of denoising steps or introducing additional inference latency. Empirically, this design improves distribution matching to the teacher, leading to more diverse outputs and better fine-grained visual quality during long-horizon rollout.

\subsubsection{GAN Control Regularization}
\label{sec:gan}

\begin{figure}[t]
    \centering
    \includegraphics[width=1\linewidth]{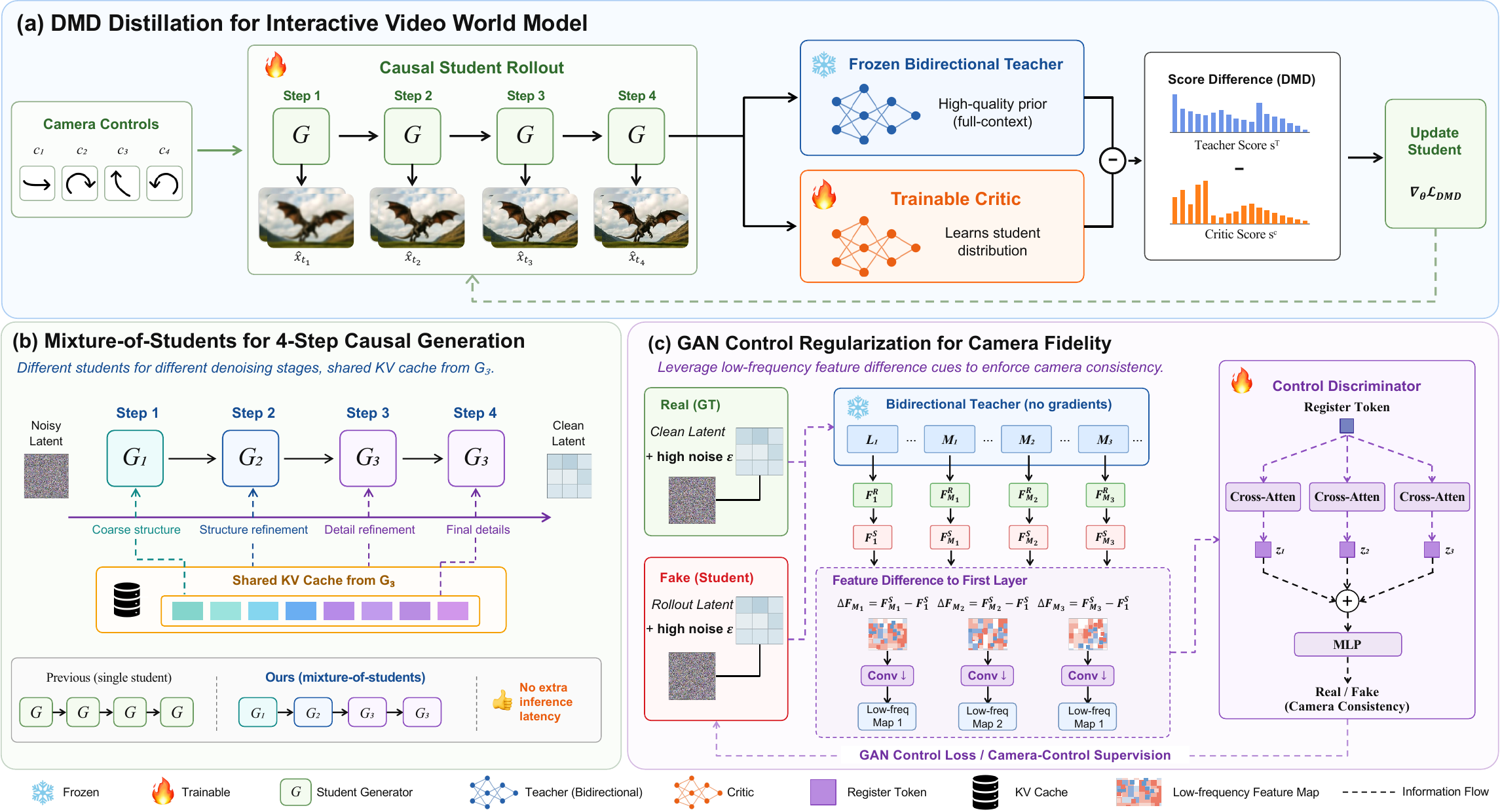}
\caption{\textbf{Distillation framework with Mixture-of-Students and GAN Control Regularization.}
(a) The causal few-step student autoregressively rolls out video chunks under camera controls. The generated rollout is perturbed with noise and evaluated by a frozen bidirectional teacher and a trainable critic, whose score difference provides the DMD gradient for updating the student. 
(b) To increase student-side capacity, we specialize different generators for different denoising stages. The 4-step sampler uses $G_1$ for coarse structure, $G_2$ for structure refinement, and $G_3$ for detail refinement and final reconstruction, while sharing the same streaming KV cache to avoid extra inference latency. 
(c) To improve camera fidelity, we introduce a control discriminator built from low-frequency teacher feature differences. Real and fake noised latents are passed through the frozen teacher with the same camera condition. Intermediate features are compared with the first-layer feature, downsampled into low-frequency maps, and aggregated by a register-token discriminator to distinguish camera-consistent real samples from student rollouts.}
    \label{fig:method}
\end{figure}

Built upon DMD~\citep{yin2024dmd1}, DMD2~\citep{yin2024dmd2} attaches an additional GAN objective to the critic and has been adopted in recent world-model distillation pipelines~\citep{gao2026advancing} to improve high-frequency details and local realism. However, these objectives are not specifically designed for camera-controlled autoregressive video generation. In our setting, we observe a distinct failure mode during DMD training: the student gradually loses camera fidelity, where the generated camera motion becomes faster, deviates from the target trajectory, and eventually leads to severe blur or scene collapse. This issue is amplified in causal rollout, since camera-control errors accumulate over time.

We attribute this camera drifting problem to insufficient control supervision in the standard DMD objective. The score residuals produced by DMD mainly emphasize high-frequency local artifacts, such as edges and textures, but provide weak gradients for the low-frequency layout changes induced by camera motion. Once the student rollout drifts far from the teacher's training distribution, e.g., when the camera has already flown away or the scene is heavily degraded, the frozen bidirectional teacher can no longer provide reliable score guidance. Prior work mitigates this instability by adding an additional downsampled $\ell_2$ loss between self-rollouts and ground-truth videos~\citep{hong2025relic}. While effective for stabilization, such direct regression tends to shrink the distribution and often produces blurry newly explored regions.

To address this issue, we introduce \emph{GAN control regularization}, an adversarial objective that focuses on low-frequency camera-consistency cues rather than high-frequency texture realism. As shown in Figure~\ref{fig:method}(c), we construct real and fake noised latents by perturbing the ground-truth latent and the student rollout latent with a high noise level:
\begin{equation}
\mathbf{z}^{r}_{\tau}
=
\alpha_{\tau}\mathbf{z}_{0}
+
\sigma_{\tau}\boldsymbol{\epsilon},
\qquad
\mathbf{z}^{f}_{\tau}
=
\alpha_{\tau}\hat{\mathbf{z}}_{0}
+
\sigma_{\tau}\boldsymbol{\epsilon},
\qquad
\boldsymbol{\epsilon}\sim\mathcal{N}(\mathbf{0},\mathbf{I}),
\end{equation}
where $\mathbf{z}_{0}$ denotes the clean ground-truth latent, $\hat{\mathbf{z}}_{0}$ denotes the student rollout latent, and $\tau$ is sampled from a high-noise range. Both latents are forwarded through the frozen camera-conditioned bidirectional teacher under the same camera condition $\mathbf{c}$. Since the teacher is fixed, its intermediate representations provide a stable camera-aware feature space for comparing real and fake trajectories. We empirically find this design more effective than extracting features from the trainable critic, whose representation keeps changing during distillation training.

Instead of discriminating the final latent directly, our control discriminator operates on teacher feature dynamics. Since the camera condition is injected at the input, the first-layer feature provides a direct camera-aware reference. We compare intermediate teacher features against this reference to assess whether the input latent remains consistent with the prescribed camera trajectory.
For $x\in\{r,f\}$, where $r$ denotes the real ground-truth latent and $f$ denotes the fake student rollout, we extract the first-layer and three selected middle-layer features from the frozen teacher:
\begin{equation}
\left(
\mathbf{F}^{x}_{1},
\mathbf{F}^{x}_{M_1},
\mathbf{F}^{x}_{M_2},
\mathbf{F}^{x}_{M_3}
\right)
=
T_{\mathrm{feat}}\!\left(\mathbf{z}^{x}_{\tau}, \mathbf{c}\right),
\end{equation}
where $T_{\mathrm{feat}}$ denotes the frozen camera-conditioned teacher feature extractor. For each middle layer, we compute a low-frequency feature-difference map with respect to the first-layer feature:
\begin{equation}
\Delta \mathbf{F}^{x}_{M_i}
=
\psi_i\!\left(
\eta_i\!\left(\mathbf{F}^{x}_{M_i}\right)
-
\eta_1\!\left(\mathbf{F}^{x}_{1}\right)
\right),
\qquad i\in\{1,2,3\}.
\end{equation}
Here, $\eta_1$ and $\eta_i$ are linear feature extractors for the first layer and layer $M_i$, and $\psi_i$ is a strided convolutional downsampler that suppresses high-frequency texture details and encourages the discriminator to focus on low-frequency layout and camera-motion consistency. We then aggregate the resulting feature-difference maps with 3 learnable register tokens. The register tokens attend to their corresponding low-frequency maps through cross-attention, and the resulting token features are summed to produce a control-consistency logit:
\begin{equation}
\mathbf{h}^{x}_{i}
=
\operatorname{CrossAttn}_{i}
\left(
\mathbf{q}, \Delta \mathbf{F}^{x}_{M_i}
\right),
\qquad
d^{x}_{\mathrm{ctrl}}
=
\operatorname{MLP}
\left(
\left[\sum_{i=1}^{3}\mathbf{h}^{x}_{i}, e_\tau \right]
\right),
\quad x\in\{r,f\}.
\end{equation}
Here, $\mathbf{q}$ is the learnable register token, $e_\tau$ refers to the timestep embedding, and $d^{r}_{\mathrm{ctrl}}$ and $d^{f}_{\mathrm{ctrl}}$ denote the discriminator logits for real ground-truth latents and fake student rollout latents, respectively. We train the control discriminator and the student generator with a relativistic softplus adversarial loss:
\begin{equation}
\mathcal{L}^{D}_{\mathrm{ctrl}}
=
\mathbb{E}_{\tau,\boldsymbol{\epsilon}}
\left[
\operatorname{softplus}
\left(
d^{f}_{\mathrm{ctrl}} - d^{r}_{\mathrm{ctrl}}
\right)
\right],
\qquad
\mathcal{L}^{G}_{\mathrm{ctrl}}
=
\mathbb{E}_{\tau,\boldsymbol{\epsilon}}
\left[
\operatorname{softplus}
\left(
d^{r}_{\mathrm{ctrl}} - d^{f}_{\mathrm{ctrl}}
\right)
\right].
\end{equation}
The generator is optimized with $\mathcal{L}_{\mathrm{DMD}}+\lambda_{\mathrm{ctrl}}\mathcal{L}^{G}_{\mathrm{ctrl}}$.
Compared with the high-frequency adversarial objective in DMD2, our control regularization provides a complementary low-frequency supervision signal for camera fidelity. Empirically, the proposed objective stabilizes the distillation process, substantially improves camera following during autoregressive rollout, and avoids the mode shrinkage introduced by prior regression-based stabilization losses.

\subsection{Inference-time Optimization}
To achieve low-latency inference, we conduct code optimizations following ~\citep{hong2025relic}. We briefly describe techniques below. Because inference latency can be caused by GPU memory traffic, kernel-launch overhead, and repeated transformer computation, we focus our optimizations on these aspects. For the repeated prediction and cache-update steps, we compile the model operations once, reuse the generated GPU kernels across runs, and replay the same CUDA execution pattern to reduce launch overhead. We adopt self-attention KV caches to preserve previously computed video context, while using text cross-attention caches to eliminate repeated computation of prompt key and value projections. For memory retrieval, we maintain a separate cache that stores compressed key summaries to support efficient retrieval of the relevant historical chunks. We further employ FlashAttention-3~\citep{Shah2024FlashAttention3FA} on Hopper GPUs, fused QKV projections, BF16 RMSNorm, cached RoPE embeddings, and timestep modulations. Finally, guided by profiling results, we manually restructure some PyTorch operations to eliminate additional runtime overhead.

During inference, we employ multi-GPU parallelism to distribute both computation and memory workloads across devices. FFN layers and cross-attention modules are distributed along the sequence dimension using sequence parallelism, whereas self-attention is partitioned across attention heads using tensor parallelism. NCCL \texttt{All-to-All} collectives convert between these two tensor layouts. This design also partitions the KV cache along the attention-head dimension, such that each GPU stores and processes only its assigned subset of heads. Auxiliary workloads, including text encoding and VAE processing, are offloaded from the primary generation GPUs, while a Tiny VAE adapted from ~\citep{shin2025motionstream} is used for efficient decoding of conditioning inputs and generated outputs.

To support minutes-long generation beyond our training horizon, we adopt a rolling/sliding-window caching scheme~\citep{xiao2024efficient}. Specifically, the KV cache is organized as a concatenation of \texttt{[sink chunks, top-k chunks, recent context chunks]}. Top-k chunks are selected dynamically by our sparse attention at each prediction step. To avoid unseen positional embeddings along the frame axis, we remap the frame indices back to the training horizon based on their relative distance. Together, these designs keep both the active context size and positional range bounded, enabling stable minute-scale rollouts without increasing per-step inference latency.

\subsection{Training Infrastructure and Model Details}

Training a bidirectional teacher together with three 14B student generators (MoS) over a 20-second horizon is highly memory-intensive. Following prior work~\citep{hong2025relic,gao2026advancing}, we combine several techniques to improve training efficiency: (1) Fully Sharded Data Parallel 2 (FSDP2)~\citep{zhao2023pytorch} to shard model parameters, gradients, and optimizer states across GPUs; (2) sequence parallelism to distribute the video-token sequence across devices; (3) tensor parallelism to partition self-attention heads and associated computations; (4) gradient checkpointing; and (5) gradient accumulation. During distribution-level distillation, we further adopt replayed back-propagation~\citep{hong2025relic} to reduce the memory required for differentiating through autoregressive rollouts. Together, these techniques allow us to train with a global batch size of 64 on 32 NVIDIA H200 GPUs. At inference time, our mixture-of-students sampler uses three generators for four denoising steps: $G_1$ is used at the first step, $G_2$ at the second step, and $G_3$ at the final two steps. We use a chunk size of two latent frames to maintain low streaming latency.

\section{Experiments}
\label{sec:experiments}

In this section, we evaluate \ours as a unified interactive video world model for both image- and video-conditioned generation. We compare it with recent advanced world models and show that \ours achieves the best overall visual quality and camera-following accuracy. Our experiments cover two settings. In the \textit{image-to-video} setting, the model is given a single image and must infer unseen regions while following user-specified camera controls. In the \textit{video-to-video} setting, the model is given an input video and must preserve its subject dynamics while enabling free camera navigation, thereby constructing a dynamic and controllable world. We first describe the benchmarks and evaluation metrics, followed by quantitative and qualitative results for both tasks.

\begin{table}[t]
\centering
\small
\setlength{\tabcolsep}{4.5pt}
\begin{tabular}{lcccccccc}
\toprule
\multirow{2}{*}{Model}
& \multicolumn{6}{c}{Visual quality $\uparrow$}
& \multicolumn{2}{c}{Action accuracy (RPE $\downarrow$)} \\
\cmidrule(lr){2-7} \cmidrule(lr){8-9}
& Avg. Score$^\dagger$
& Imaging
& Aesthetic
& Dynamic
& Motion
& Flickering
& Trans.
& Rot. \\
\midrule
\multicolumn{9}{l}{\textit{Image-to-Video (I2V)}} \\
HY-WorldPlay 1.5 
& 0.7900 & 0.6825 & \textbf{0.6915} & 0.6120 & \textbf{0.9893} & \textbf{0.9745} & 0.0195 & 0.1711 \\
RELIC
& 0.8225 & 0.5983 & 0.6334 & 0.9243 & 0.9889 & 0.9675 & 0.0208 & 0.1426 \\
LingBot-World-Fast
& 0.8261 & 0.6553 & 0.6360 & 0.8991 & 0.9769 & 0.9633 & 0.0174 & 0.2922 \\
SANA-WM-Streaming
& 0.8415 & 0.6480 & 0.6099 & \textbf{1.0000} & 0.9889 & 0.9608 & 0.0240 & 0.1155 \\
DreamX-World
& 0.8385 & 0.6891 & 0.6655 & 0.8927 & 0.9847 & 0.9604 & 0.0244 & 0.2547 \\
\textbf{\ours}
& \textbf{0.8558} & \textbf{0.7113} & 0.6416 & 0.9874 & 0.9794 & 0.9622
& \textbf{0.0132} & \textbf{0.0784} \\
\midrule
\multicolumn{9}{l}{\textit{Video-to-Video (V2V)}} \\
Inspatio-World
& 0.8374     & 0.6756 & 0.5815 & \textbf{1.0000} & 0.9794 & 0.9506 & 0.0436 & 0.2470  \\
\textbf{\ours}
& \textbf{0.8527} & \textbf{0.6981} & \textbf{0.6196} & \textbf{1.0000} & \textbf{0.9855} & \textbf{0.9602} & \textbf{0.0187} & \textbf{0.1119} \\
\bottomrule
\end{tabular}
\caption{
\textbf{Comparison on visual quality and action accuracy.}
$^\dagger$Average Score is the arithmetic mean of Imaging, Aesthetic, Dynamic, Motion, and Flickering scores.
We report results under both image-to-video (I2V) and video-to-video (V2V) settings. \ours achieves the best overall video synthesis quality and camera-following accuracy.
}
\label{tab:vbench_action_accuracy}
\end{table}

\begin{figure}
    \centering
    \includegraphics[width=1\linewidth]{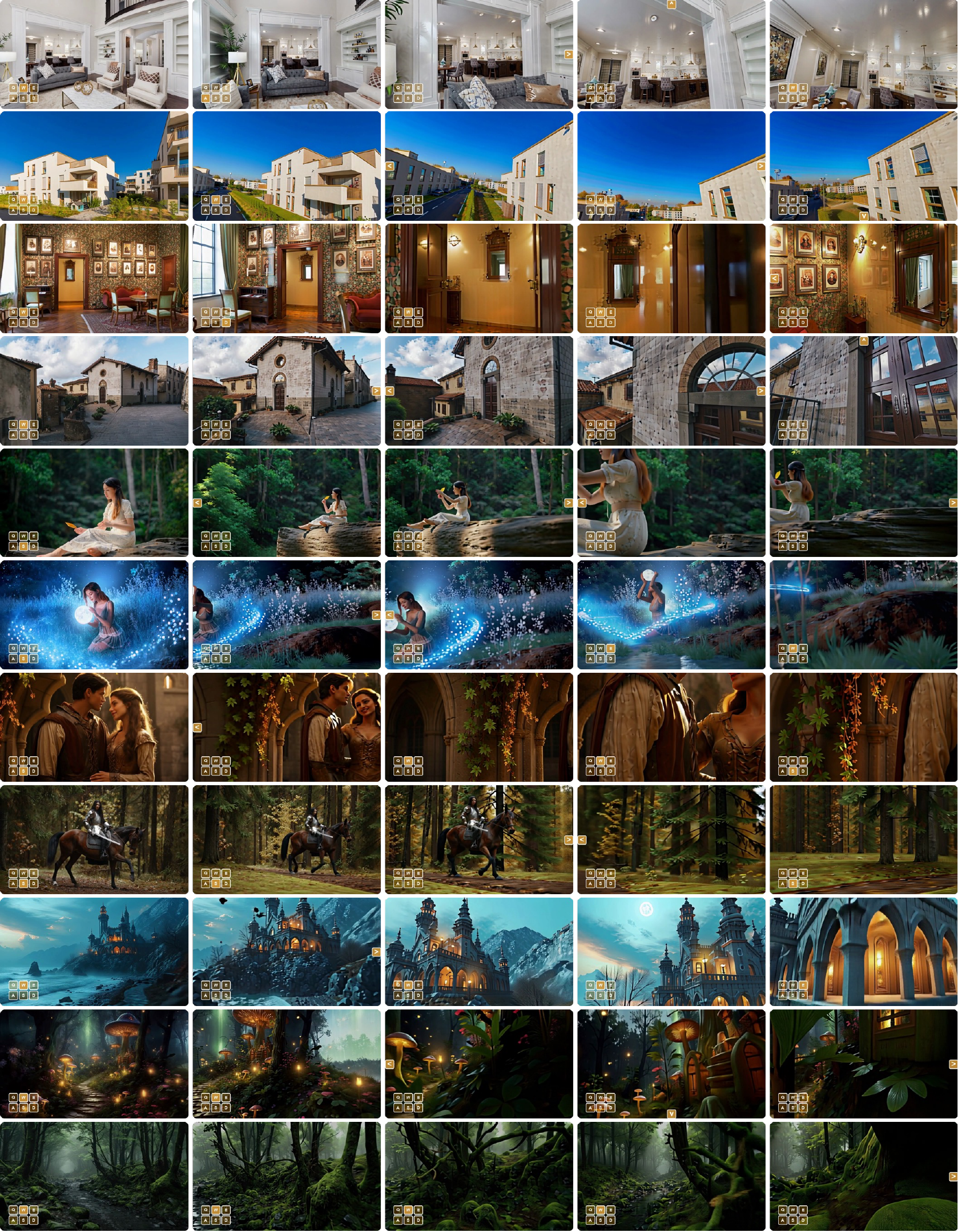}
    \caption{
    \textbf{Navigation into diverse images.}
    }
    \label{fig:demo_i2v_main}
\end{figure}

\begin{figure}
    \centering
    \includegraphics[width=1\linewidth]{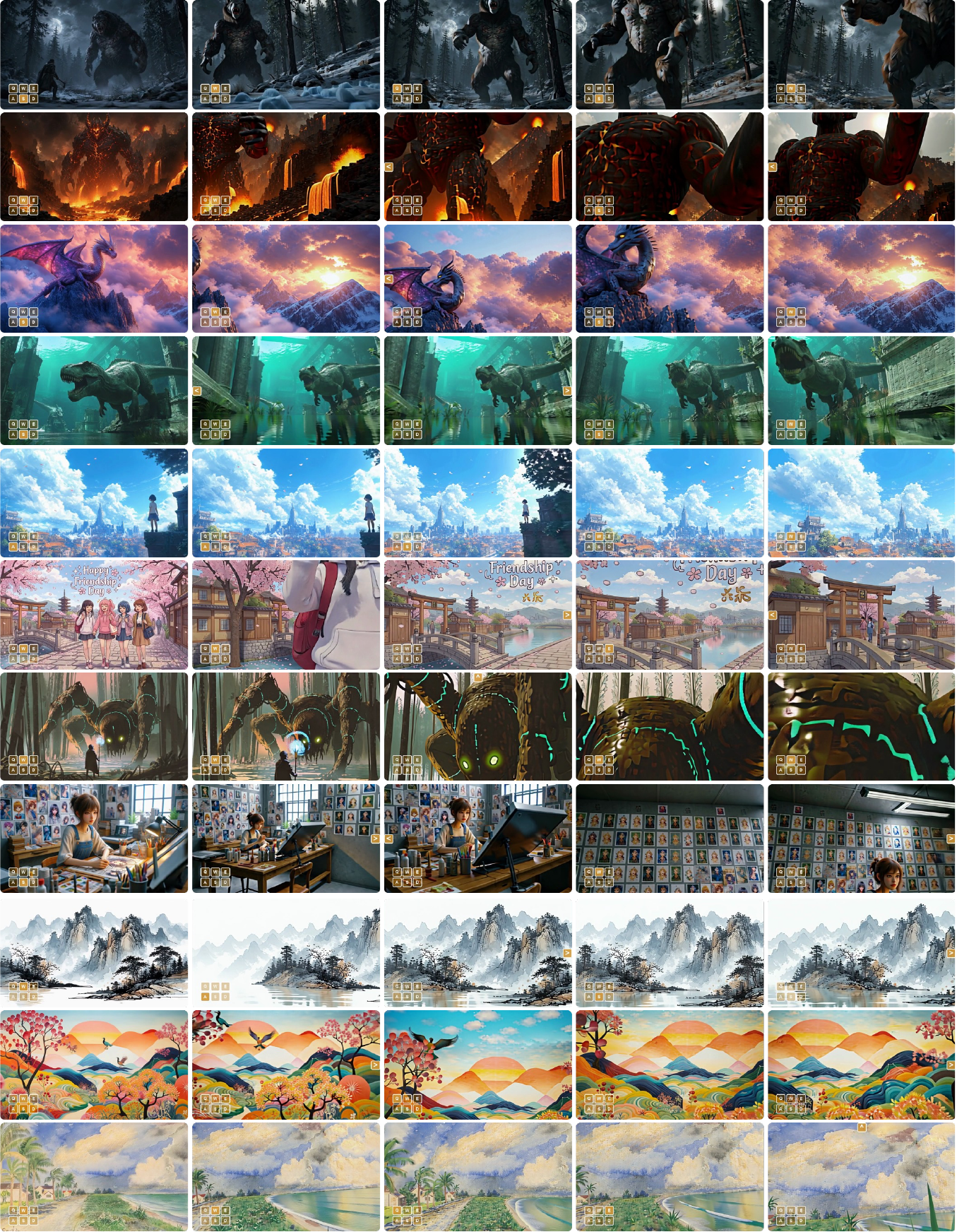}
    \caption{
    \textbf{Navigation into diverse images.}
    }
    \label{fig:demo_i2v_other}
\end{figure}

\subsection{Image-to-Video World Modeling}

\textbf{Benchmark.}
We construct a general image-to-video world-modeling benchmark with 1,000 diverse open-source images, covering real-world photographs, AI-generated images, cartoons, artistic images, and gaming scenes. For each image, we use a VLM to generate a scene caption and produce navigation keys that are consistent with the image content. Each image is paired with five camera trajectories of varying translation scales and rotation speeds, allowing us to evaluate the robustness of different models under various camera-control patterns. 

\textbf{Metrics.}
Following prior work~\citep{gao2026advancing,zhu2026sana}, we report five visual-quality metrics from VBench~\citep{huang2023vbench}: imaging quality, aesthetic quality, dynamic degree, motion smoothness, and temporal flickering. To evaluate camera-following accuracy, we estimate relative camera poses from each generated video using two pose-estimation models, Depth Anything 3~\citep{lin2025depth} and ViPE~\citep{huang2025vipe}, and average their pose errors for a more robust assessment. Since different models may produce trajectories with different translation scales and coordinate frames, we align each reconstructed trajectory to the canonical target trajectory using Umeyama alignment~\citep{umeyama2002least}. We then compute translational and rotational Relative Pose Error (RPE) to measure how accurately each model follows the prescribed camera motion.

\textbf{Baselines.}
We compare \ours with five recent interactive video world models available at the time of this report: HY-WorldPlay 1.5~\citep{team2025hunyuanworld}, RELIC~\citep{hong2025relic}, LingBot-World-Fast~\citep{gao2026advancing}, SANA-WM-Streaming~\citep{zhu2026sana}, and DreamX-World~\citep{team2026dreamx}. These models all target real-time or streaming video generation with interactive camera control, making them the most relevant baselines for our setting. For each baseline, we use the officially released model or the recommended inference setting whenever available.

\begin{figure}
    \centering
    \includegraphics[width=1\linewidth]{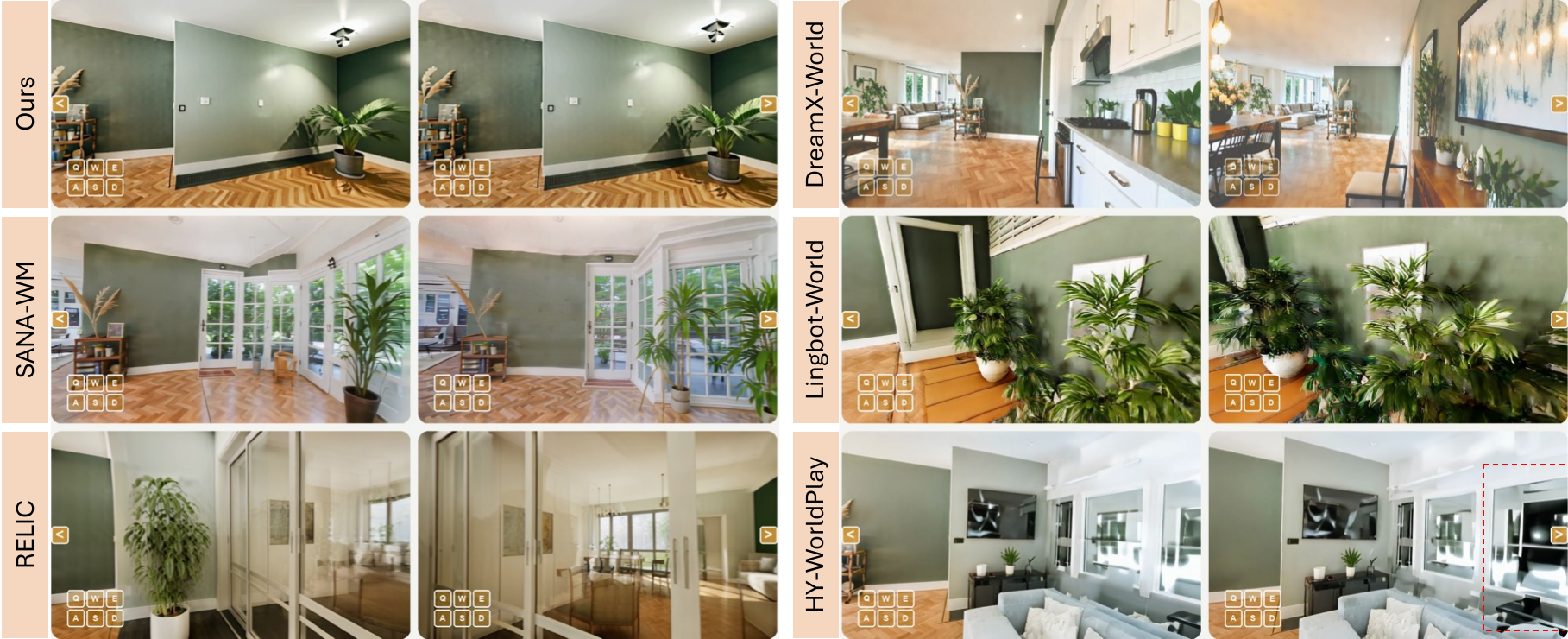}
    \caption{
    \textbf{Visual comparison of Image-to-Video world modeling performance.} We evaluate the memory capability of each model with a revisit trajectory. Specifically, the camera first pans right to observe a target region, then pans left to move away, and finally pans right again to revisit the same region. We compare the two frames captured at the same camera position to assess whether the scene content is consistently preserved. Most prior models struggle to retrieve long-term visual memory, leading to inconsistent structures or changed appearances after revisiting. RELIC~\citep{hong2025relic} and LingBot-World~\citep{gao2026advancing} further suffer from degraded visual quality under long-horizon rollout, highlighting the difficulty of maintaining both memory consistency and generation quality over extended interactive exploration.
    }
    \label{fig:i2v_compare}
\end{figure}

\textbf{Results.}
Table~\ref{tab:vbench_action_accuracy} reports the quantitative comparison under the image-to-video setting. \ours achieves the best overall visual quality with an average score of $0.8558$, outperforming all recent streaming world-model baselines. The gain mainly comes from improved imaging quality, where \ours obtains the highest score of $0.7113$, indicating stronger scene synthesis and fewer visual artifacts. Although some baselines achieve higher scores on individual temporal metrics such as motion smoothness or flickering, they often trade off camera controllability. In terms of action accuracy, \ours achieves the lowest translational and rotational RPE among all methods. Compared with the strongest baseline, \ours reduces translational error from $0.0174$ to $0.0132$ and rotational error from $0.1155$ to $0.0784$. These results demonstrate that our semi-explicit camera representation and distillation strategy substantially improve camera following while maintaining high visual quality, enabling more reliable interactive world exploration. Figure~\ref{fig:i2v_compare} presents a qualitative comparison of long-horizon memory and visual quality between \ours and recent state-of-the-art models. Most prior methods fail to faithfully reconstruct previously observed regions upon revisit, while others suffer from substantial visual degradation as the rollout length increases.

We present more qualitative results of \ours in Figure~\ref{fig:demo_i2v_main} and Figure~\ref{fig:demo_i2v_other}. Across diverse input images, \ours can understand and preserve the underlying scene structure while synthesizing plausible unseen regions under user-controlled camera motion. The generated views follow the prescribed camera instructions accurately, and newly discovered areas remain semantically consistent with the input image while exhibiting high visual quality and diverse content. Moreover, when the camera revisits previously observed regions after long-horizon exploration, \ours preserves scene details and appearance consistency, demonstrating the effectiveness of our sparse memory mechanism. Benefiting from multi-task training, \ours also exhibits dynamic world-modeling capability in the I2V setting, producing meaningful object and subject motion that is consistent with the scene semantics.

\subsection{Video-to-Video World Modeling}

\textbf{Benchmark \& Metrics.}
Similar to the I2V setting, we construct a video-to-video benchmark with 500 videos covering diverse scenes. Unlike the I2V benchmark, all videos contain dynamic subjects, such as humans, animals, and vehicles, allowing us to evaluate whether a model can preserve source-video dynamics while supporting camera-controlled world exploration. We use a VLM to generate video captions and user navigation keys for each input video. Each video is paired with 6 camera trajectories, which are grouped into three types: \textit{retreat}, where the camera moves backward to reveal the full object or scene; \textit{follow}, where the camera tracks the moving subject and keeps it visible; and \textit{free}, where the camera performs unconstrained exploratory navigation. We use the same visual-quality and camera-following metrics as in the I2V setting. Specifically, VBench metrics are used to evaluate generation quality, while translational and rotational RPE are computed from estimated camera trajectories to measure action-following accuracy.

\textbf{Baselines.}
We omit short-horizon bidirectional re-shotting models~\citep{bai2025recammaster,huang2025spacetimepilot,wang2026bullettime}, as they are not designed for long-video generation or real-time streaming. At the time of this report, the only open-source baseline that supports long-horizon V2V world modeling is Inspatio-World~\citep{team2026inspatio}. Inspatio-World adopts point-cloud re-rendering as its camera-control signal and uses a sliding-window strategy by discarding distant video tokens for efficient inference. However, as discussed in previous sections, point-cloud re-rendering is inherently limited by the reconstructed view coverage and is therefore less suitable for open-ended free exploration. Meanwhile, discarding distant tokens removes long-term visual memory, making it difficult to preserve consistency when revisiting previously observed regions. Notably, unlike \textcolor{FutureOrange}{\textbf{Wonder}}, Inspatio-World is not designed for streaming inference, as its point-cloud re-rendering pipeline requires access to the entire input video.
\begin{figure}
    \centering
    \includegraphics[width=1\linewidth]{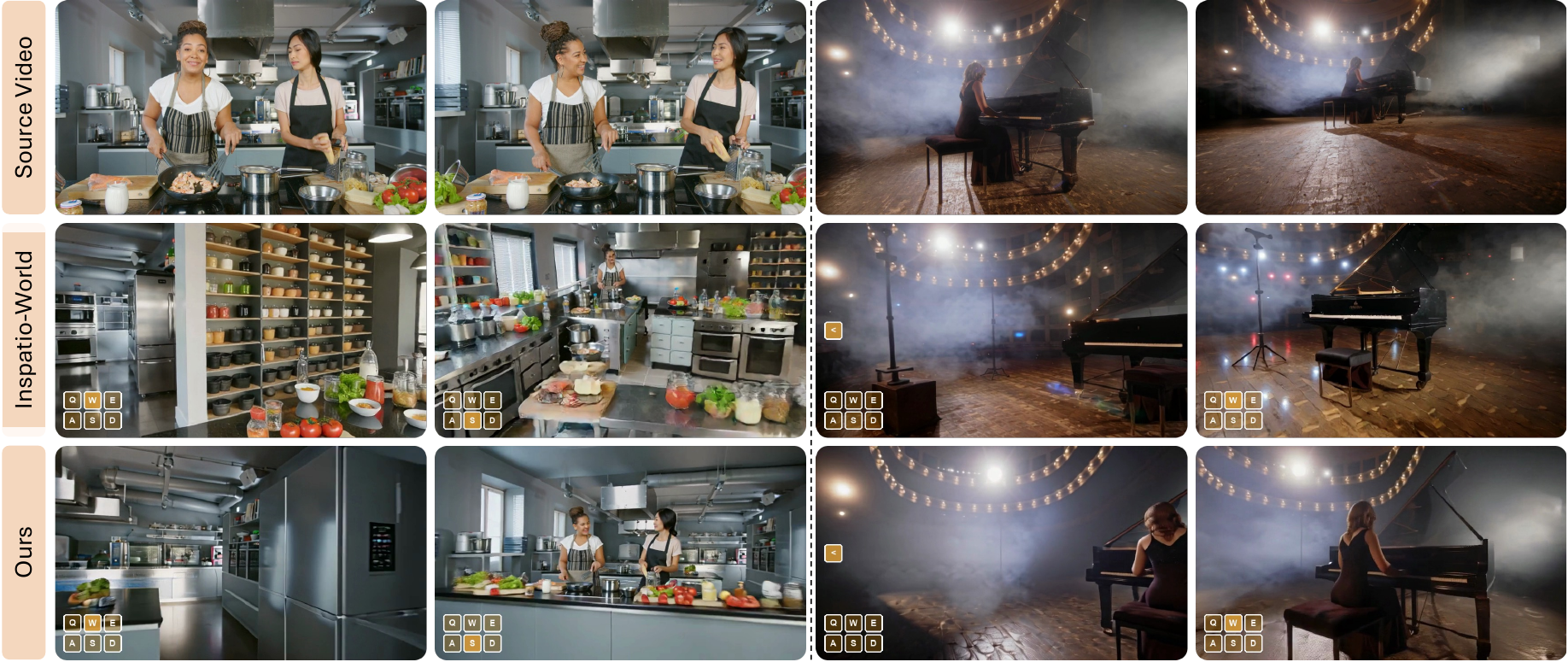}
    \caption{
    \textbf{Visual comparison of Video-to-Video world modeling performance.} Due to errors in the point clouds reconstructed by external pose-estimation models and the limited exploration space of point-cloud representations, Inspatio-World~\citep{team2026inspatio} often struggles to produce plausible camera motion and to preserve dynamic objects under large viewpoint changes.
    }
    \label{fig:v2v_compare}
\end{figure}

\begin{figure}
    \centering
    \includegraphics[width=1\linewidth]{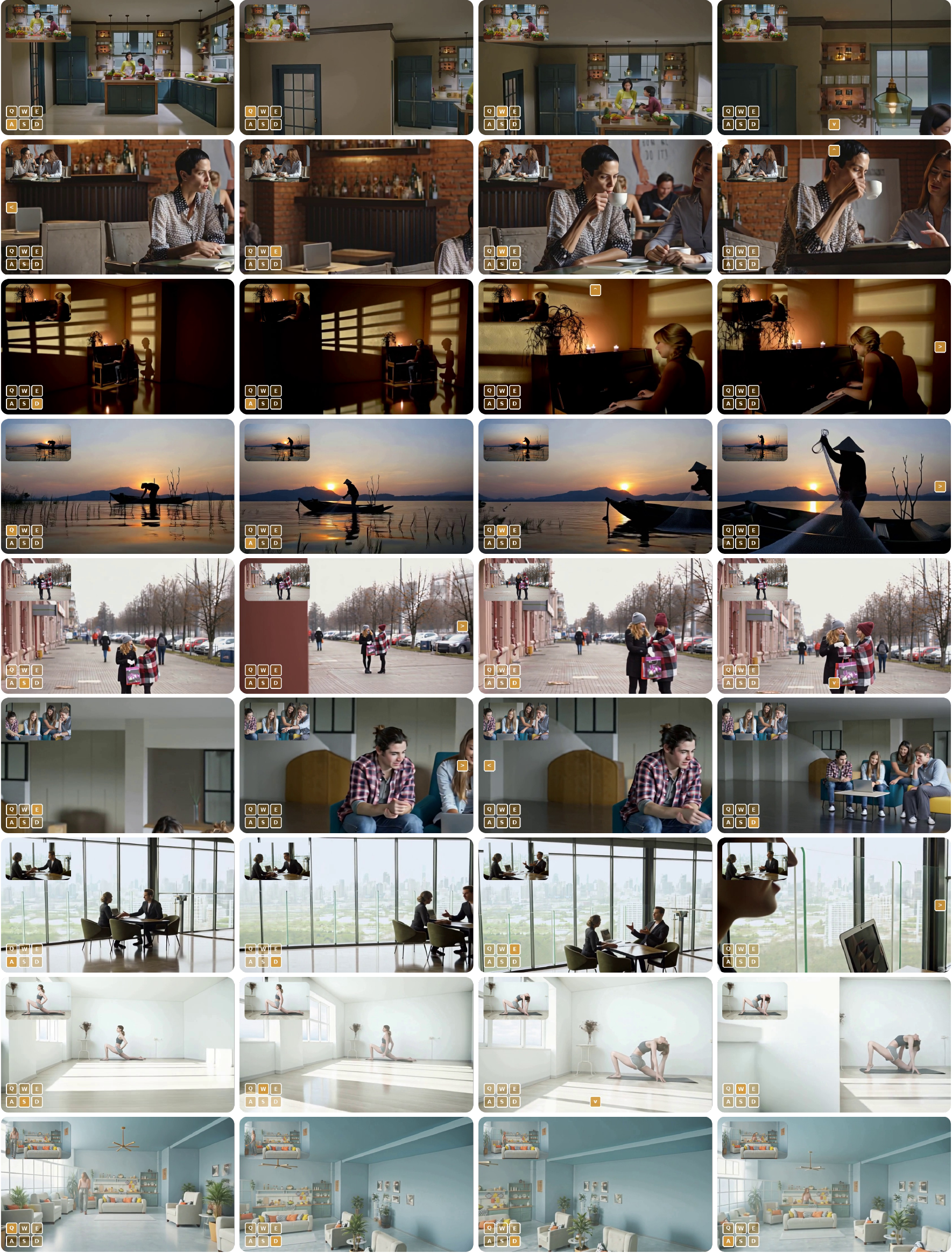}
    \caption{
    \textbf{Navigation into videos.}
    }
    \label{fig:demo_v2v_main}
\end{figure}

\textbf{Results.}
Table~\ref{tab:vbench_action_accuracy} reports the quantitative results under the video-to-video setting. Compared with Inspatio-World, \ours achieves better overall visual quality, improving the average score from $0.8374$ to $0.8527$. The gains are consistent across most visual metrics, including imaging quality, aesthetic quality, motion smoothness, and temporal flickering. This indicates that \ours can better preserve the dynamic content of the input video while synthesizing high-quality target views under user-controlled camera motion. More importantly, \ours substantially improves camera-following accuracy. It reduces translational RPE from $0.0436$ to $0.0187$ and rotational RPE from $0.2470$ to $0.1119$, showing much stronger alignment with the prescribed camera trajectories. These improvements demonstrate the advantage of our semi-explicit lattice control over point-cloud re-rendering, especially when the camera moves beyond the reconstructed view coverage. The qualitative results shown in Figure~\ref{fig:v2v_compare} and Figure~\ref{fig:demo_v2v_main} further demonstrate that \ours enables streaming V2V world modeling with better dynamic preservation, stronger camera following, and more stable long-horizon visual consistency. We believe \ours also establishes a strong baseline for video re-shotting.

\section{Conclusion}
\label{sec:conclusion}

We introduced \textcolor{FutureOrange}{\textbf{Wonder}}, a real-time camera-controllable video world model that transforms a single image into an explorable world and re-renders an observed video along user-specified camera trajectories while preserving its appearance and dynamics. \ours jointly addresses controllability, persistence, and efficiency through a unified system design. A dense pixel-space coordinate field provides spatially aligned camera cues, a sparse full-fidelity memory mechanism supports coherent revisits with bounded active attention cost, and an improved autoregressive distillation pipeline preserves visual quality and camera fidelity in a few-step causal student. Across both image- and video-conditioned benchmarks, \ours achieves stronger visual quality and camera-following accuracy than recent streaming world models, while preserving dynamic content under novel viewpoints and maintaining scene consistency over long rollouts. These advances enable minute-scale generation at 16 FPS with stable inference latency, unifying open-ended exploration beyond observed views and real-time re-rendering of dynamic scenes within a single model.

\clearpage
\newpage
\bibliographystyle{assets/plainnat}
\bibliography{paper}

\begin{thebibliography}{51}
\providecommand{\natexlab}[1]{#1}
\providecommand{\url}[1]{\texttt{#1}}
\expandafter\ifx\csname urlstyle\endcsname\relax
  \providecommand{\doi}[1]{doi: #1}\else
  \providecommand{\doi}{doi: \begingroup \urlstyle{rm}\Url}\fi

\bibitem[Bai et~al.(2025{\natexlab{a}})Bai, Xia, Fu, Wang, Mu, Cao, Liu, Hu, Bai, Wan, et~al.]{bai2025recammaster}
Jianhong Bai, Menghan Xia, Xiao Fu, Xintao Wang, Lianrui Mu, Jinwen Cao, Zuozhu Liu, Haoji Hu, Xiang Bai, Pengfei Wan, et~al.
\newblock Recammaster: Camera-controlled generative rendering from a single video.
\newblock In \emph{Proceedings of the IEEE/CVF International Conference on Computer Vision}, pages 14834--14844, 2025{\natexlab{a}}.

\bibitem[Bai et~al.(2025{\natexlab{b}})Bai, Cai, Chen, Chen, Chen, Cheng, Deng, Ding, Gao, Ge, et~al.]{bai2025qwen3}
Shuai Bai, Yuxuan Cai, Ruizhe Chen, Keqin Chen, Xionghui Chen, Zesen Cheng, Lianghao Deng, Wei Ding, Chang Gao, Chunjiang Ge, et~al.
\newblock Qwen3-vl technical report.
\newblock \emph{arXiv preprint arXiv:2511.21631}, 2025{\natexlab{b}}.

\bibitem[Ball et~al.(2025)Ball, Bauer, Belletti, Brownfield, Ephrat, Fruchter, Gupta, Holsheimer, Holynski, Hron, Kaplanis, Limont, McGill, Oliveira, Parker-Holder, Perbet, Scully, Shar, Spencer, Tov, Villegas, Wang, Yung, Baetu, Berbel, Bridson, Bruce, Buttimore, Chakera, Chandra, Collins, Cullum, Damoc, Dasagi, Gazeau, Gbadamosi, Han, Hirst, Kachra, Kerley, Kjems, Knoepfel, Koriakin, Lo, Lu, Mehring, Moufarek, Nandwani, Oliveira, Pardo, Park, Pierson, Poole, Ran, Salimans, Sanchez, Saprykin, Shen, Sidhwani, Smith, Stanton, Tomlinson, Vijaykumar, Wang, Wingfield, Wong, Xu, Yew, Young, Zubov, Eck, Erhan, Kavukcuoglu, Hassabis, Gharamani, Hadsell, van~den Oord, Mosseri, Bolton, Singh, and Rockt{\"a}schel]{genie3}
Philip~J. Ball, Jakob Bauer, Frank Belletti, Bethanie Brownfield, Ariel Ephrat, Shlomi Fruchter, Agrim Gupta, Kristian Holsheimer, Aleksander Holynski, Jiri Hron, Christos Kaplanis, Marjorie Limont, Matt McGill, Yanko Oliveira, Jack Parker-Holder, Frank Perbet, Guy Scully, Jeremy Shar, Stephen Spencer, Omer Tov, Ruben Villegas, Emma Wang, Jessica Yung, Cip Baetu, Jordi Berbel, David Bridson, Jake Bruce, Gavin Buttimore, Sarah Chakera, Bilva Chandra, Paul Collins, Alex Cullum, Bogdan Damoc, Vibha Dasagi, Maxime Gazeau, Charles Gbadamosi, Woohyun Han, Ed~Hirst, Ashyana Kachra, Lucie Kerley, Kristian Kjems, Eva Knoepfel, Vika Koriakin, Jessica Lo, Cong Lu, Zeb Mehring, Alex Moufarek, Henna Nandwani, Valeria Oliveira, Fabio Pardo, Jane Park, Andrew Pierson, Ben Poole, Helen Ran, Tim Salimans, Manuel Sanchez, Igor Saprykin, Amy Shen, Sailesh Sidhwani, Duncan Smith, Joe Stanton, Hamish Tomlinson, Dimple Vijaykumar, Luyu Wang, Piers Wingfield, Nat Wong, Keyang Xu, Christopher Yew, Nick Young, Vadim Zubov, Douglas
  Eck, Dumitru Erhan, Koray Kavukcuoglu, Demis Hassabis, Zoubin Gharamani, Raia Hadsell, A{\"a}ron van~den Oord, Inbar Mosseri, Adrian Bolton, Satinder Singh, and Tim Rockt{\"a}schel.
\newblock Genie 3: A new frontier for world models, 2025.
\newblock \url{https://deepmind.google/discover/blog/genie-3-a-new-frontier-for-world-models/}.

\bibitem[Chen et~al.(2024)Chen, Mart{\'\i}~Mons{\'o}, Du, Simchowitz, Tedrake, and Sitzmann]{chen2024dforcing}
Boyuan Chen, Diego Mart{\'\i}~Mons{\'o}, Yilun Du, Max Simchowitz, Russ Tedrake, and Vincent Sitzmann.
\newblock Diffusion forcing: Next-token prediction meets full-sequence diffusion.
\newblock \emph{Advances in Neural Information Processing Systems}, 37:\penalty0 24081--24125, 2024.

\bibitem[Chung et~al.(2023)Chung, Constant, Garc{\'i}a, Roberts, Tay, Narang, and Firat]{Chung2023UniMaxFA}
Hyung~Won Chung, Noah Constant, Xavier Garc{\'i}a, Adam Roberts, Yi~Tay, Sharan Narang, and Orhan Firat.
\newblock Unimax: Fairer and more effective language sampling for large-scale multilingual pretraining.
\newblock \emph{ArXiv}, abs/2304.09151, 2023.
\newblock \url{https://api.semanticscholar.org/CorpusID:258187051}.

\bibitem[Cui et~al.(2025)Cui, Wu, Li, Yang, Li, Wang, Bai, Ban, and Hsieh]{cui2025self}
Justin Cui, Jie Wu, Ming Li, Tao Yang, Xiaojie Li, Rui Wang, Andrew Bai, Yuanhao Ban, and Cho-Jui Hsieh.
\newblock Self-forcing++: Towards minute-scale high-quality video generation.
\newblock \emph{arXiv preprint arXiv:2510.02283}, 2025.

\bibitem[Gao et~al.(2024)Gao, Shi, Zhang, Wang, Xiao, and Chen]{gao2024ca2}
Kaifeng Gao, Jiaxin Shi, Hanwang Zhang, Chunping Wang, Jun Xiao, and Long Chen.
\newblock Ca2-vdm: Efficient autoregressive video diffusion model with causal generation and cache sharing.
\newblock \emph{arXiv preprint arXiv:2411.16375}, 2024.

\bibitem[Guo et~al.(2025)Guo, Ye, He, Wu, Jiang, Pearce, and Bian]{guo2025mineworld}
Junliang Guo, Yang Ye, Tianyu He, Haoyu Wu, Yushu Jiang, Tim Pearce, and Jiang Bian.
\newblock Mineworld: a real-time and open-source interactive world model on minecraft.
\newblock \emph{arXiv preprint arXiv:2504.08388}, 2025.

\bibitem[HaCohen et~al.(2026)HaCohen, Brazowski, Chiprut, Bitterman, Kvochko, Berkowitz, Shalem, Lifschitz, Moshe, Porat, et~al.]{hacohen2026ltx}
Yoav HaCohen, Benny Brazowski, Nisan Chiprut, Yaki Bitterman, Andrew Kvochko, Avishai Berkowitz, Daniel Shalem, Daphna Lifschitz, Dudu Moshe, Eitan Porat, et~al.
\newblock Ltx-2: Efficient joint audio-visual foundation model.
\newblock \emph{arXiv preprint arXiv:2601.03233}, 2026.

\bibitem[He et~al.(2025{\natexlab{a}})He, Xu, Guo, Wetzstein, Dai, Li, and Yang]{he2025cameractrl}
Hao He, Yinghao Xu, Yuwei Guo, Gordon Wetzstein, Bo~Dai, Hongsheng Li, and Ceyuan Yang.
\newblock Cameractrl: Enabling camera control for video diffusion models.
\newblock In \emph{The Thirteenth International Conference on Learning Representations}, 2025{\natexlab{a}}.

\bibitem[He et~al.(2025{\natexlab{b}})He, Peng, Liu, Wang, Zhang, Cui, Kang, Jiang, An, Ren, et~al.]{he2025matrixgame2}
Xianglong He, Chunli Peng, Zexiang Liu, Boyang Wang, Yifan Zhang, Qi~Cui, Fei Kang, Biao Jiang, Mengyin An, Yangyang Ren, et~al.
\newblock Matrix-game 2.0: An open-source, real-time, and streaming interactive world model.
\newblock \emph{arXiv preprint arXiv:2508.13009}, 2025{\natexlab{b}}.

\bibitem[Hong et~al.(2025)Hong, Mei, Ge, Xu, Zhou, Bi, Hold-Geoffroy, Roberts, Fisher, Shechtman, et~al.]{hong2025relic}
Yicong Hong, Yiqun Mei, Chongjian Ge, Yiran Xu, Yang Zhou, Sai Bi, Yannick Hold-Geoffroy, Mike Roberts, Matthew Fisher, Eli Shechtman, et~al.
\newblock Relic: Interactive video world model with long-horizon memory.
\newblock \emph{arXiv preprint arXiv:2512.04040}, 2025.

\bibitem[Hu et~al.(2024)Hu, Hu, Song, Huang, Wang, Zhou, Liu, Ma, and Sun]{hu2024acdit}
Jinyi Hu, Shengding Hu, Yuxuan Song, Yufei Huang, Mingxuan Wang, Hao Zhou, Zhiyuan Liu, Wei-Ying Ma, and Maosong Sun.
\newblock Acdit: Interpolating autoregressive conditional modeling and diffusion transformer.
\newblock \emph{arXiv preprint arXiv:2412.07720}, 2024.

\bibitem[Huang et~al.(2025{\natexlab{a}})Huang, Zhou, Rabeti, Korovko, Ling, Ren, Shen, Gao, Slepichev, Lin, Ren, Xie, Biswas, Leal-Taixe, and Fidler]{huang2025vipe}
Jiahui Huang, Qunjie Zhou, Hesam Rabeti, Aleksandr Korovko, Huan Ling, Xuanchi Ren, Tianchang Shen, Jun Gao, Dmitry Slepichev, Chen-Hsuan Lin, Jiawei Ren, Kevin Xie, Joydeep Biswas, Laura Leal-Taixe, and Sanja Fidler.
\newblock Vipe: Video pose engine for 3d geometric perception.
\newblock In \emph{NVIDIA Research Whitepapers}, 2025{\natexlab{a}}.

\bibitem[Huang et~al.(2025{\natexlab{b}})Huang, Hu, Han, Shi, Tian, He, and Jiang]{huang2025memory}
Junchao Huang, Xinting Hu, Boyao Han, Shaoshuai Shi, Zhuotao Tian, Tianyu He, and Li~Jiang.
\newblock Memory forcing: Spatio-temporal memory for consistent scene generation on minecraft.
\newblock \emph{arXiv preprint arXiv:2510.03198}, 2025{\natexlab{b}}.

\bibitem[Huang et~al.(2025{\natexlab{c}})Huang, Li, He, Zhou, and Shechtman]{huang2025self}
Xun Huang, Zhengqi Li, Guande He, Mingyuan Zhou, and Eli Shechtman.
\newblock Self forcing: Bridging the train-test gap in autoregressive video diffusion.
\newblock In \emph{Advances in neural information processing systems}, 2025{\natexlab{c}}.

\bibitem[Huang et~al.(2025{\natexlab{d}})Huang, Jeong, Chen, Gryaditskaya, Wang, Lasenby, and Huang]{huang2025spacetimepilot}
Zhening Huang, Hyeonho Jeong, Xuelin Chen, Yulia Gryaditskaya, Tuanfeng~Y Wang, Joan Lasenby, and Chun-Hao Huang.
\newblock Spacetimepilot: Generative rendering of dynamic scenes across space and time.
\newblock \emph{arXiv preprint arXiv:2512.25075}, 2025{\natexlab{d}}.

\bibitem[Huang et~al.(2024)Huang, He, Yu, Zhang, Si, Jiang, Zhang, Wu, Jin, Chanpaisit, Wang, Chen, Wang, Lin, Qiao, and Liu]{huang2023vbench}
Ziqi Huang, Yinan He, Jiashuo Yu, Fan Zhang, Chenyang Si, Yuming Jiang, Yuanhan Zhang, Tianxing Wu, Qingyang Jin, Nattapol Chanpaisit, Yaohui Wang, Xinyuan Chen, Limin Wang, Dahua Lin, Yu~Qiao, and Ziwei Liu.
\newblock {VBench}: Comprehensive benchmark suite for video generative models.
\newblock In \emph{Proceedings of the IEEE/CVF Conference on Computer Vision and Pattern Recognition}, 2024.

\bibitem[Kong et~al.(2024)Kong, Tian, Zhang, Min, Dai, Zhou, Xiong, Li, Wu, Zhang, et~al.]{kong2024hunyuanvideo}
Weijie Kong, Qi~Tian, Zijian Zhang, Rox Min, Zuozhuo Dai, Jin Zhou, Jiangfeng Xiong, Xin Li, Bo~Wu, Jianwei Zhang, et~al.
\newblock Hunyuanvideo: A systematic framework for large video generative models.
\newblock \emph{arXiv preprint arXiv:2412.03603}, 2024.

\bibitem[Li et~al.(2025)Li, Tang, Xu, Wu, Zhou, Shao, Yu, Cao, and Lu]{li2025hunyuangame}
Jiaqi Li, Junshu Tang, Zhiyong Xu, Longhuang Wu, Yuan Zhou, Shuai Shao, Tianbao Yu, Zhiguo Cao, and Qinglin Lu.
\newblock Hunyuan-gamecraft: High-dynamic interactive game video generation with hybrid history condition.
\newblock \emph{arXiv preprint arXiv:2506.17201}, 2025.

\bibitem[Lin et~al.(2025)Lin, Chen, Liew, Chen, Li, Shi, Feng, and Kang]{lin2025depth}
Haotong Lin, Sili Chen, Junhao Liew, Donny~Y Chen, Zhenyu Li, Guang Shi, Jiashi Feng, and Bingyi Kang.
\newblock Depth anything 3: Recovering the visual space from any views.
\newblock \emph{arXiv preprint arXiv:2511.10647}, 2025.

\bibitem[Ling et~al.(2024)Ling, Sheng, Tu, Zhao, Xin, Wan, Yu, Guo, Yu, Lu, et~al.]{ling2024dl3dv}
Lu~Ling, Yichen Sheng, Zhi Tu, Wentian Zhao, Cheng Xin, Kun Wan, Lantao Yu, Qianyu Guo, Zixun Yu, Yawen Lu, et~al.
\newblock Dl3dv-10k: A large-scale scene dataset for deep learning-based 3d vision.
\newblock In \emph{Proceedings of the IEEE/CVF Conference on Computer Vision and Pattern Recognition}, pages 22160--22169, 2024.

\bibitem[Liu et~al.(2025)Liu, Hu, Xu, Shan, and Lu]{liu2025rolling}
Kunhao Liu, Wenbo Hu, Jiale Xu, Ying Shan, and Shijian Lu.
\newblock Rolling forcing: Autoregressive long video diffusion in real time.
\newblock \emph{arXiv preprint arXiv:2509.25161}, 2025.

\bibitem[Mao et~al.(2025)Mao, Lin, Li, Li, Peng, He, Pang, Chi, Qiao, and Zhang]{mao2025yume}
Xiaofeng Mao, Shaoheng Lin, Zhen Li, Chuanhao Li, Wenshuo Peng, Tong He, Jiangmiao Pang, Mingmin Chi, Yu~Qiao, and Kaipeng Zhang.
\newblock Yume: An interactive world generation model.
\newblock \emph{arXiv preprint arXiv:2507.17744}, 2025.

\bibitem[Peng et~al.(2023)Peng, Quesnelle, Fan, and Shippole]{peng2023yarn}
Bowen Peng, Jeffrey Quesnelle, Honglu Fan, and Enrico Shippole.
\newblock Yarn: Efficient context window extension of large language models.
\newblock \emph{arXiv preprint arXiv:2309.00071}, 2023.

\bibitem[Rahimi et~al.(2025)Rahimi, Sadeghi-Niaraki, and Choi]{rahimi2025generative}
Fatema Rahimi, Abolghasem Sadeghi-Niaraki, and Soo-Mi Choi.
\newblock Generative ai meets virtual reality: A comprehensive survey on applications, challenges, and future direction.
\newblock \emph{IEEE Access}, 2025.

\bibitem[{Robbyant Team}(2026)]{gao2026advancing}
{Robbyant Team}.
\newblock Advancing open-source world models.
\newblock \emph{arXiv preprint arXiv:2601.20540}, 2026.

\bibitem[Shah et~al.(2024)Shah, Bikshandi, Zhang, Thakkar, Ramani, and Dao]{Shah2024FlashAttention3FA}
Jay Shah, Ganesh Bikshandi, Ying Zhang, Vijay Thakkar, Pradeep Ramani, and Tri Dao.
\newblock Flashattention-3: Fast and accurate attention with asynchrony and low-precision.
\newblock \emph{ArXiv}, abs/2407.08608, 2024.
\newblock \url{https://api.semanticscholar.org/CorpusID:271098045}.

\bibitem[Shin et~al.(2025)Shin, Li, Zhang, Zhu, Park, Schechtman, and Huang]{shin2025motionstream}
Joonghyuk Shin, Zhengqi Li, Richard Zhang, Jun-Yan Zhu, Jaesik Park, Eli Schechtman, and Xun Huang.
\newblock Motionstream: Real-time video generation with interactive motion controls.
\newblock \emph{arXiv preprint arXiv:2511.01266}, 2025.

\bibitem[Tang et~al.(2025)Tang, Liu, Li, Wu, Yang, Zhao, Gong, Yuan, Shao, Zhang, et~al.]{tang2025hunyuan2}
Junshu Tang, Jiacheng Liu, Jiaqi Li, Longhuang Wu, Haoyu Yang, Penghao Zhao, Siruis Gong, Xiang Yuan, Shuai Shao, Linfeng Zhang, et~al.
\newblock Hunyuan-gamecraft-2: Instruction-following interactive game world model.
\newblock \emph{arXiv preprint arXiv:2511.23429}, 2025.

\bibitem[Team et~al.(2026{\natexlab{a}})Team, Bai, Chen, Chu, Dang, Dou, Gao, Gu, Hong, Lei, et~al.]{team2026dreamx}
DreamX Team, Yancheng Bai, Rui Chen, Xiangxiang Chu, Rujing Dang, Hao Dou, Bingjie Gao, Qiwen Gu, Siyu Hong, Jiachen Lei, et~al.
\newblock Dreamx-world 1.0: A general-purpose interactive world model.
\newblock \emph{arXiv preprint arXiv:2606.16993}, 2026{\natexlab{a}}.

\bibitem[Team et~al.(2025)Team, Wang, Liu, Wu, Gu, Wang, Zuo, Huang, Li, Zhang, et~al.]{team2025hunyuanworld}
HunyuanWorld Team, Zhenwei Wang, Yuhao Liu, Junta Wu, Zixiao Gu, Haoyuan Wang, Xuhui Zuo, Tianyu Huang, Wenhuan Li, Sheng Zhang, et~al.
\newblock Hunyuanworld 1.0: Generating immersive, explorable, and interactive 3d worlds from words or pixels.
\newblock \emph{arXiv preprint arXiv:2507.21809}, 2025.

\bibitem[Team et~al.(2026{\natexlab{b}})Team, Shen, Zhang, Liu, Ji, Bao, Zhai, Liu, Guo, Wang, et~al.]{team2026inspatio}
InSpatio Team, Donghui Shen, Guofeng Zhang, Haomin Liu, Haoyu Ji, Hujun Bao, Hongjia Zhai, Jialin Liu, Jing Guo, Nan Wang, et~al.
\newblock Inspatio-world: A real-time 4d world simulator via spatiotemporal autoregressive modeling.
\newblock \emph{arXiv preprint arXiv:2604.07209}, 2026{\natexlab{b}}.

\bibitem[Umeyama(2002)]{umeyama2002least}
Shinji Umeyama.
\newblock Least-squares estimation of transformation parameters between two point patterns.
\newblock \emph{IEEE Transactions on pattern analysis and machine intelligence}, 13\penalty0 (4):\penalty0 376--380, 2002.

\bibitem[Wan et~al.(2025)Wan, Wang, Ai, Wen, Mao, Xie, Chen, Yu, Zhao, Yang, et~al.]{wan2025wan}
Team Wan, Ang Wang, Baole Ai, Bin Wen, Chaojie Mao, Chen-Wei Xie, Di~Chen, Feiwu Yu, Haiming Zhao, Jianxiao Yang, et~al.
\newblock Wan: Open and advanced large-scale video generative models.
\newblock \emph{arXiv preprint arXiv:2503.20314}, 2025.

\bibitem[Wang et~al.(2026{\natexlab{a}})Wang, Zhang, Cai, Wu, Ackermann, Kuang, Zheng, Raji{\v{c}}, Tang, and Wetzstein]{wang2026bullettime}
Yiming Wang, Qihang Zhang, Shengqu Cai, Tong Wu, Jan Ackermann, Zhengfei Kuang, Yang Zheng, Frano Raji{\v{c}}, Siyu Tang, and Gordon Wetzstein.
\newblock Bullettime: Decoupled control of time and camera pose for video generation.
\newblock In \emph{Proceedings of the IEEE/CVF Conference on Computer Vision and Pattern Recognition}, pages 18319--18330, 2026{\natexlab{a}}.

\bibitem[Wang et~al.(2024)Wang, Yuan, Wang, Li, Chen, Xia, Luo, and Shan]{wang2024motionctrl}
Zhouxia Wang, Ziyang Yuan, Xintao Wang, Yaowei Li, Tianshui Chen, Menghan Xia, Ping Luo, and Ying Shan.
\newblock Motionctrl: A unified and flexible motion controller for video generation.
\newblock In \emph{ACM SIGGRAPH 2024 Conference Papers}, pages 1--11, 2024.

\bibitem[Wang et~al.(2026{\natexlab{b}})Wang, Liu, Li, Huang, Xu, Kang, An, Wang, Jiang, Wei, et~al.]{wang2026matrix3}
Zile Wang, Zexiang Liu, Jiaxing Li, Kaichen Huang, Baixin Xu, Fei Kang, Mengyin An, Peiyu Wang, Biao Jiang, Yichen Wei, et~al.
\newblock Matrix-game 3.0: Real-time and streaming interactive world model with long-horizon memory.
\newblock \emph{arXiv preprint arXiv:2604.08995}, 2026{\natexlab{b}}.

\bibitem[Xiao et~al.(2024)Xiao, Tian, Chen, Han, and Lewis]{xiao2024efficient}
Guangxuan Xiao, Yuandong Tian, Beidi Chen, Song Han, and Mike Lewis.
\newblock Efficient streaming language models with attention sinks.
\newblock In \emph{International Conference on Learning Representations}, volume 2024, pages 21875--21895, 2024.

\bibitem[Xiao et~al.(2025)Xiao, Yang, Chang, Chen, Xiong, Xu, Zheng, and Zhang]{xiao2025world}
Junjin Xiao, Yandan Yang, Xinyuan Chang, Ronghan Chen, Feng Xiong, Mu~Xu, Wei-Shi Zheng, and Qing Zhang.
\newblock World-env: Leveraging world model as a virtual environment for vla post-training.
\newblock \emph{arXiv preprint arXiv:2509.24948}, 2025.

\bibitem[Xu et~al.(2025)Xu, Mei, Zhang, and Patel]{xu2025freevis}
Jiacong Xu, Yiqun Mei, Ke~Zhang, and Vishal~M Patel.
\newblock Freevis: Training-free video stylization with inconsistent references.
\newblock \emph{arXiv preprint arXiv:2510.01686}, 2025.

\bibitem[Yang et~al.(2025)Yang, Teng, Zheng, Ding, Huang, Xu, Yang, Hong, Zhang, Feng, et~al.]{yang2025cogvideox}
Zhuoyi Yang, Jiayan Teng, Wendi Zheng, Ming Ding, Shiyu Huang, Jiazheng Xu, Yuanming Yang, Wenyi Hong, Xiaohan Zhang, Guanyu Feng, et~al.
\newblock Cogvideox: Text-to-video diffusion models with an expert transformer.
\newblock In \emph{International Conference on Learning Representations}, volume 2025, pages 83048--83077, 2025.

\bibitem[Yin et~al.(2024{\natexlab{a}})Yin, Gharbi, Park, Zhang, Shechtman, Durand, and Freeman]{yin2024dmd2}
Tianwei Yin, Micha{\"e}l Gharbi, Taesung Park, Richard Zhang, Eli Shechtman, Fredo Durand, and Bill Freeman.
\newblock Improved distribution matching distillation for fast image synthesis.
\newblock \emph{Advances in neural information processing systems}, 37:\penalty0 47455--47487, 2024{\natexlab{a}}.

\bibitem[Yin et~al.(2024{\natexlab{b}})Yin, Gharbi, Zhang, Shechtman, Durand, Freeman, and Park]{yin2024dmd1}
Tianwei Yin, Micha{\"e}l Gharbi, Richard Zhang, Eli Shechtman, Fredo Durand, William~T Freeman, and Taesung Park.
\newblock One-step diffusion with distribution matching distillation.
\newblock In \emph{Proceedings of the IEEE/CVF conference on computer vision and pattern recognition}, pages 6613--6623, 2024{\natexlab{b}}.

\bibitem[Yin et~al.(2025)Yin, Zhang, Zhang, Freeman, Durand, Shechtman, and Huang]{yin2025causvid}
Tianwei Yin, Qiang Zhang, Richard Zhang, William~T Freeman, Fredo Durand, Eli Shechtman, and Xun Huang.
\newblock From slow bidirectional to fast autoregressive video diffusion models.
\newblock In \emph{Proceedings of the Computer Vision and Pattern Recognition Conference}, pages 22963--22974, 2025.

\bibitem[Yu et~al.(2025)Yu, Bai, Qin, Liu, Wang, Wan, Zhang, and Liu]{yu2025contextasmem}
Jiwen Yu, Jianhong Bai, Yiran Qin, Quande Liu, Xintao Wang, Pengfei Wan, Di~Zhang, and Xihui Liu.
\newblock Context as memory: Scene-consistent interactive long video generation with memory retrieval.
\newblock \emph{arXiv preprint arXiv:2506.03141}, 2025.

\bibitem[Yu et~al.(2024)Yu, Xing, Yuan, Hu, Li, Huang, Gao, Wong, Shan, and Tian]{yu2024viewcrafter}
Wangbo Yu, Jinbo Xing, Li~Yuan, Wenbo Hu, Xiaoyu Li, Zhipeng Huang, Xiangjun Gao, Tien-Tsin Wong, Ying Shan, and Yonghong Tian.
\newblock Viewcrafter: Taming video diffusion models for high-fidelity novel view synthesis.
\newblock \emph{arXiv preprint arXiv:2409.02048}, 2024.

\bibitem[Zhang et~al.(2025)Zhang, Peng, Wang, Wang, Zhu, Kang, Jiang, Gao, Li, Liu, et~al.]{zhang2025matrixgame1}
Yifan Zhang, Chunli Peng, Boyang Wang, Puyi Wang, Qingcheng Zhu, Fei Kang, Biao Jiang, Zedong Gao, Eric Li, Yang Liu, et~al.
\newblock Matrix-game: Interactive world foundation model.
\newblock \emph{arXiv preprint arXiv:2506.18701}, 2025.

\bibitem[Zhao et~al.(2023)Zhao, Gu, Varma, Luo, Huang, Xu, Wright, Shojanazeri, Ott, Shleifer, et~al.]{zhao2023pytorch}
Yanli Zhao, Andrew Gu, Rohan Varma, Liang Luo, Chien-Chin Huang, Min Xu, Less Wright, Hamid Shojanazeri, Myle Ott, Sam Shleifer, et~al.
\newblock Pytorch fsdp: experiences on scaling fully sharded data parallel.
\newblock \emph{arXiv preprint arXiv:2304.11277}, 2023.

\bibitem[Zhu et~al.(2026{\natexlab{a}})Zhu, Liu, Zhao, Ye, Chen, Yu, He, Han, and Xie]{zhu2026sana}
Haoyi Zhu, Haozhe Liu, Yuyang Zhao, Tian Ye, Junsong Chen, Jincheng Yu, Tong He, Song Han, and Enze Xie.
\newblock Sana-wm: Efficient minute-scale world modeling with hybrid linear diffusion transformer.
\newblock \emph{arXiv preprint arXiv:2605.15178}, 2026{\natexlab{a}}.

\bibitem[Zhu et~al.(2026{\natexlab{b}})Zhu, Zhao, He, Su, Li, and Zhu]{zhu2026causal}
Hongzhou Zhu, Min Zhao, Guande He, Hang Su, Chongxuan Li, and Jun Zhu.
\newblock Causal forcing: Autoregressive diffusion distillation done right for high-quality real-time interactive video generation.
\newblock \emph{arXiv preprint arXiv:2602.02214}, 2026{\natexlab{b}}.

\end{thebibliography}

\end{document}